%% file: anonymous-submission-latex-2026.tex
\documentclass[letterpaper]{article} 
\usepackage[draft]{aaai2026}  
\usepackage{times}  
\usepackage{helvet}  
\usepackage{courier}  
\usepackage[hyphens]{url}  
\usepackage{graphicx} 
\usepackage{natbib}  
\usepackage{caption} 
\usepackage{algorithm}
\usepackage{algorithmic}

\usepackage{newfloat}
\usepackage{listings}
\usepackage{cleveref}
\usepackage{booktabs}
\DeclareCaptionStyle{ruled}{labelfont=normalfont,labelsep=colon,strut=off} 
\floatstyle{ruled}
\newfloat{listing}{tb}{lst}{}
\floatname{listing}{Listing}
\title{What Makes a Good Generated Image? \\ Investigating Human and Multimodal LLM Image Preference Alignment}
\author{
    Rishab Parthasarathy\textsuperscript{\rm 1}, Jasmine Collins\textsuperscript{\rm 2}, Cory Stephenson\textsuperscript{\rm 2}
}
\affiliations{
    \textsuperscript{\rm 1}MIT, \textsuperscript{\rm 2}Databricks Mosaic AI \\
    rpartha@mit.edu, jasmine.collins@databricks.com, cory.stephenson@databricks.com
}

\begin{document}
\maketitle
\begin{figure*}
\centering
\includegraphics[width=0.9\linewidth]{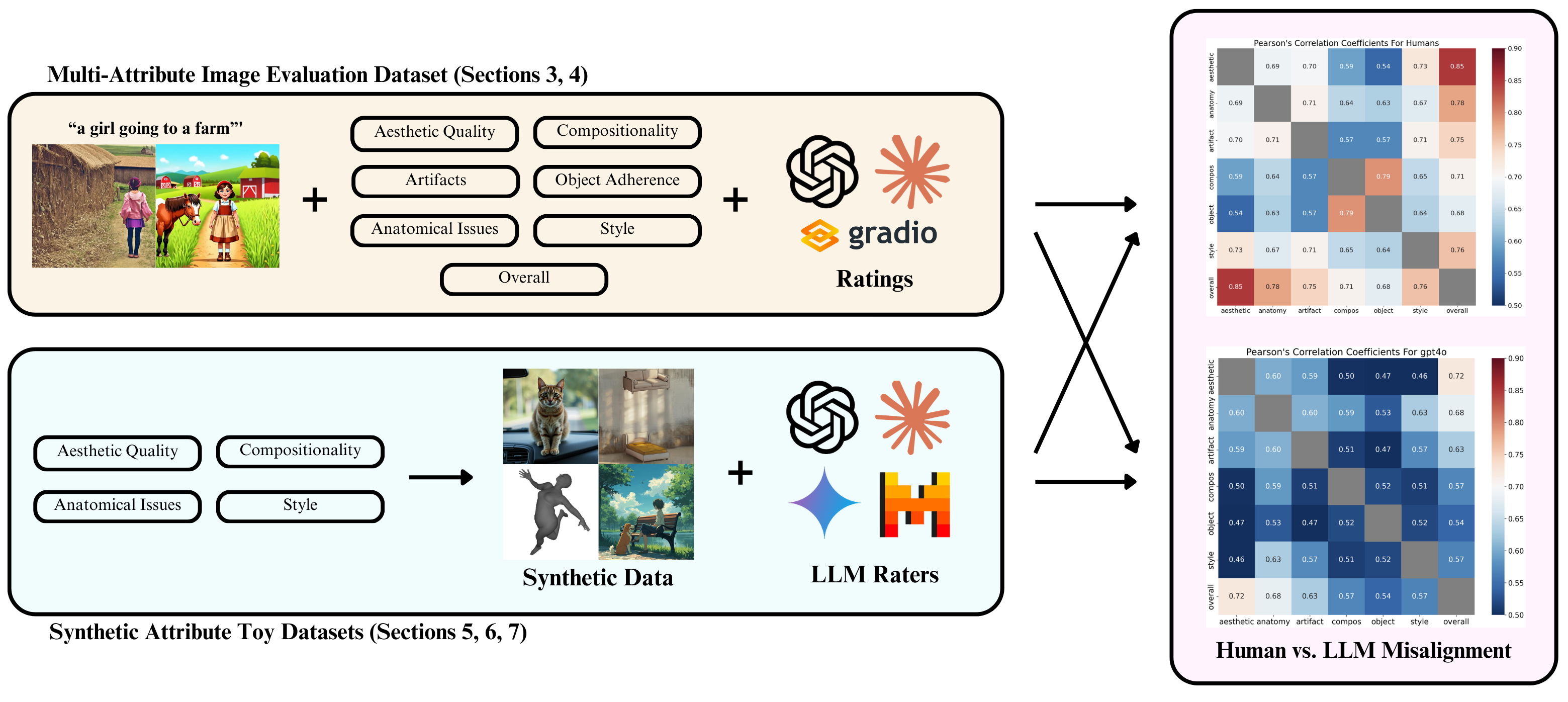}
\caption{We collect pairwise preference data across multiple axes of image quality and compute correlations between different axes of image quality. We also generate synthetic toy datasets for four of the image quality axes. Our approach enables us to determine which image quality attributes are most relevant to how humans and multimodal LLMs judge images, as well as which quality attributes are most difficult for LLMs to learn and judge, determining alignment gaps between human and LLMs.}
\label{fig:1} 
\end{figure*}

\begin{abstract}
Automated evaluation of generative text-to-image models remains a challenging problem. Recent works have proposed using multimodal LLMs to judge the quality of images, but these works offer little insight into how multimodal LLMs make use of concepts relevant to humans, such as image style or composition, to generate their overall assessment. In this work, we study what attributes of an image--specifically aesthetics, lack of artifacts, anatomical accuracy, compositional correctness, object adherence, and style--are important for both LLMs and humans to make judgments on image quality. We first curate a dataset of human preferences using synthetically generated image pairs. We use inter-task correlation between each pair of image quality attributes to understand which attributes are related in making human judgments. Repeating the same analysis with LLMs, we find that the relationships between image quality attributes are much weaker. Finally, we study individual image quality attributes by generating synthetic datasets with a high degree of control for each axis. Humans are able to easily judge the quality of an image with respect to all of the specific image quality attributes (e.g. high vs. low aesthetic image), however we find that some attributes, such as anatomical accuracy, are much more difficult for multimodal LLMs to learn to judge. Taken together, these findings reveal interesting differences between how humans and multimodal LLMs perceive images.

\end{abstract}

\section{Introduction}

Recent advancements in diffusion models have led to the rapid spread of text-to-image models, such as Stable Diffusion~\cite{rombach2021high, podell2023sdxl} and FLUX~\cite{flux1dev2024}, which have far greater capabilities than image generation models of the past. However, these more advanced models have failure modes that can be difficult to detect during evaluation. In some cases, the images may contain artifacts, not follow the prompt precisely, or may not suit the aesthetic desires of the user~\cite{cao2024synartifact, jiao2024imgdiff,huang2024aesexpert,zhou2024uniaa}. To this end, recent work has increasingly focused on improving evaluation, which can be divided into two categories: human and automated~\cite{lee2023imageeval,zheng2023llmasajudge}. Although human evaluation is still considered the gold standard, undertaking human studies has become challenging with the rapid speed of model iteration.

As a result, recent works prioritize automated evaluation, often through LLM-as-a-Judge, where a large pretrained or finetuned language model is used to judge the quality of model outputs as a proxy for human preferences~\cite{zheng2023llmasajudge}. In the case of evaluating generative diffusion models, this process typically involves either prompting a small contrastively trained multimodal LLM or a large frontier LLM with the generated image, and using the model to provide judgments on the quality of outputs. While these approaches have great potential for evaluating diffusion-generated images, recent works focus primarily on their alignment with humans on separately evaluated tasks like bias, prompt alignment, safety, and perceptual quality, with little focus on the interaction of these tasks in producing an overall image judgment~\cite{chen2024mjbench, wu2024visionprefer, huang2023t2icompbench, lee2024prometheusvision}.

In this work, we study which aspects of an image are important to humans and LLMs for judging pairwise image quality. Specifically, we curate a dataset of pairs of images generated by open weight diffusion models, which are then evaluated by human raters on six attributes typically implicated in the human evaluation of images: aesthetic quality, artifacts, anatomical issues, correctness of compositional relationships, object adherence, and style. Since this dataset uniquely has human ratings on all of these attributes for the same image pairs, we can construct inter-attribute correlations for both LLMs and humans, allowing us to study whether LLMs and humans share a standardized paradigm for evaluating images. We find that even though LLMs may seem to have outputs aligned with humans, their internal image evaluation metrics may not.

We then generate synthetic datasets based on four of our image quality attributes--aesthetic, anatomy, compositionality, and style--to further study which aspects of image evaluation are more or less challenging to multimodal LLMs. Each of these tasks involves a single-image task instead of pairwise comparison, where the LLM is either prompted to output a Yes/No rating or a numerical rating from 1-10. Using these synthetic single-axis datasets, we find that certain tasks like evaluating compositional correctness and style prove easy for LLMs to handle, but a significant gap still remains in judging anatomy validity, where humans outperform LLMs.

In summary, we investigate whether multimodal LLMs share the same attribute values as humans when evaluating diffusion-generated images using a custom curated dataset where each image pair is rated by both multimodal LLMs and humans along seven different axes. We also conduct a fine grained evaluation of these same frontier LLMs on toy versions of the tasks evaluated before, comparing them to small finetuned models and human raters. \newline\newline \textbf{Our key findings are as follows}:
\begin{itemize}
    \item LLMs and finetuned reward models perform similarly as overall image judges, and align well with humans.
    \item LLMs and humans show significant differences in how specific image attributes influence their overall rating.
    \item LLMs struggle to generalize when evaluated on basic tasks including judging anatomical correctness, image style, and aesthetics, which are easy for humans.
\end{itemize}
These findings imply that even though small models and frontier LLMs can match human performance on the overall image rating task, they process the contents of the image quite differently, and can fail to generalize in surprising ways. Our work suggests that despite impressive overall benchmark performance, there is further work to be done in aligning multimodal LLMs with human preferences.
\section{Related Work}
\label{Background}

\subsection{Multimodal Modeling}
Recent research in multimodality has focused on adding visual input to transformer models originally designed for language. These models can be fully pretrained from scratch on interleaved multimodal data, like Chameleon~\cite{chameleon2024, bai2023qwenvl}, or take pretrained LLMs and add additional modalities by training vision adapters, like LLaVA~\cite{liu2023visual, liu2023improved} and Flamingo~\cite{alayrac2022flamingo}. These vision adapters transform visual image inputs into the embedding space of the pretrained LLM, often using a small contrastive vision-language model (VLM) like CLIP~\cite{dai2024instructblip, radford2021learning, li2022blip}. In this work, due to compute and data constraints, we focus on frontier multimodal LLMs and LLaVA-based models that we train with a LLaMA-3-8B backbone~\cite{llama3}.
\subsection{Reward Modeling and Image Generation}
Current works on generative models focus on aligning them with human preferences, which directly matches real world use cases. Reinforcement learning-based approaches like RLHF and DDPO have shown that with adequate reward signal, both LLMs and generative text-to-image models can be tuned to align with arbitrary human preferences, highlighting the need for powerful reward modeling~\cite{ouyang2022rlhf, black2024ddpo, wallace2024diffusiondpo, lambert2024rewardbench, yuan2024selfrewarding, wang2024rewardmodeling, stiennon2020summarize}. 

The first class of such reward models are based off of backbones like CLIP or BLIP, which are pretrained using a contrastive training loss. Here, groups of labeled images are ingested simultaneously to align the model along a general multimodal reward axis~\cite{li2022blip,radford2021learning, xu2023imagereward}. Finetuned reward models like ImageReward~\cite{xu2023imagereward} and PickScore~\cite{kirstain2023pick} instead focus on the task of human preferences, and finetune BLIP~\cite{li2022blip} and CLIP~\cite{radford2021learning} on curated datasets of image pairs, labeled with human preferences. These classifiers lose the general image labeling capabilities of the BLIP/CLIP backbones, but are found to be significantly more aligned with human preferences~\cite{xu2023imagereward, xu2024visionreward, kirstain2023pick, wang2024incrementalreward, dzabraev2024vlrm}. On top of that, with recent advancements in scaling multimodal LLMs~\cite{bai2023qwenvl, llama3}, recent research has also focused using large frontier LLMs as generalist reward models, where a finetuning-free approach allows for plug-and-play usage with various evaluation and reinforcement learning pipelines~\cite{jiao2024imgdiff, huang2023t2icompbench, cao2024synartifact}. This pipeline typically involves using large models like GPT-4V as a rater that iteratively improves generated data, such as in the GORS finetuning pipeline from T2I-CompBench~\cite{huang2023t2icompbench}.

\subsection{LLM-as-a-Judge}
In sync with research on frontier LLMs as reward models, recent work has proposed a new evaluation paradigm termed ``LLM-as-a-Judge". Due to the high costs of human evaluation, LLM-as-a-Judge provides a platform for rapid evaluation of subjective tasks, using a frontier LLM as a proxy for human preferences. In NLP, LLM-as-a-Judge has seen rapid adoption in platforms like MT-Bench~\cite{zheng2023llmasajudge} because frontier LLMs are highly aligned with human preferences on text~\cite{son2024llmlimitations, wei2024systematicllmasjudge}. However, research in multimodal LLM-as-a-Judge for generated images remains a more open area, as multimodal frontier models are relatively recent advancements. The majority of multimodal LLM evaluations still focus on generating text, while a few works like MJ-Bench and VisionPrefer instead focus on directly augmenting diffusion model capability~\cite{chen2024mjbench,wu2024visionprefer,chen2024mllmasjudge, wang2025mllmassafety}. These existing pairwise evaluations investigate a number of different image axes, but none contain multiple ratings and evaluations for the same pair of images~\cite{kirstain2023pick, jiao2024imgdiff, chen2024mjbench, huang2024aesexpert, huang2023t2icompbench}. Instead, they feature different challenging image pairs for each task, such as safety, image difference captioning, or image quality rating~\cite{wang2025mllmassafety, chen2024mllmasjudge, jiao2024imgdiff, zhang2025qeval100kevaluatingvisualquality, han2024evalmuse40kreliablefinegrainedbenchmark}. While these datasets can thus provide insight into individual tasks, they struggle to provide insight into how LLM decisions are made on the same set of images. This paper hence focuses on a small set of axes, which are all evaluated on the same set of images, to evaluate how the learned preferences of multimodal LLMs align with humans in the task of image evaluation, and highlighting where multimodal LLMs lag behind human capabilities.

\section{Human Curated Evaluation Dataset}
\label{multi-task-dataset}
\label{sub:dataset-annotation}

We curate an evaluation dataset where each data point has seven human-generated annotations: aesthetic quality, presence of artifacts, issues with anatomy, compositional correctness, object adherence, style, and overall rating. Compositional correctness refers to whether the image correctly describes compositional aspects in the prompt like direction and location of objects. Each image pair is sampled using a prompt from PartiPrompts and generated from a sweep of 14 top open models from the ImgSys leaderboard (see Appendix Section 1 for the list of models and generation settings)~\cite{yu2023parti, imgsys2025}. These images are provided to 40+ human raters (AI researchers who rated 30+ images each) pairwise in a UI (detailed in Appendix Section 1)  where users are asked to rate the images on each of the discussed rating axes using a five-level Likert scale. For special cases, a ``not relevant'' rating is also provided. A sample set of ratings is provided in Fig. \ref{fig:2}. We provide additional example ratings in Appendix Section 3 and specific task definitions in Appendix Section 2.

\begin{figure}[t!]
    \centering
    \includegraphics[width=\linewidth]{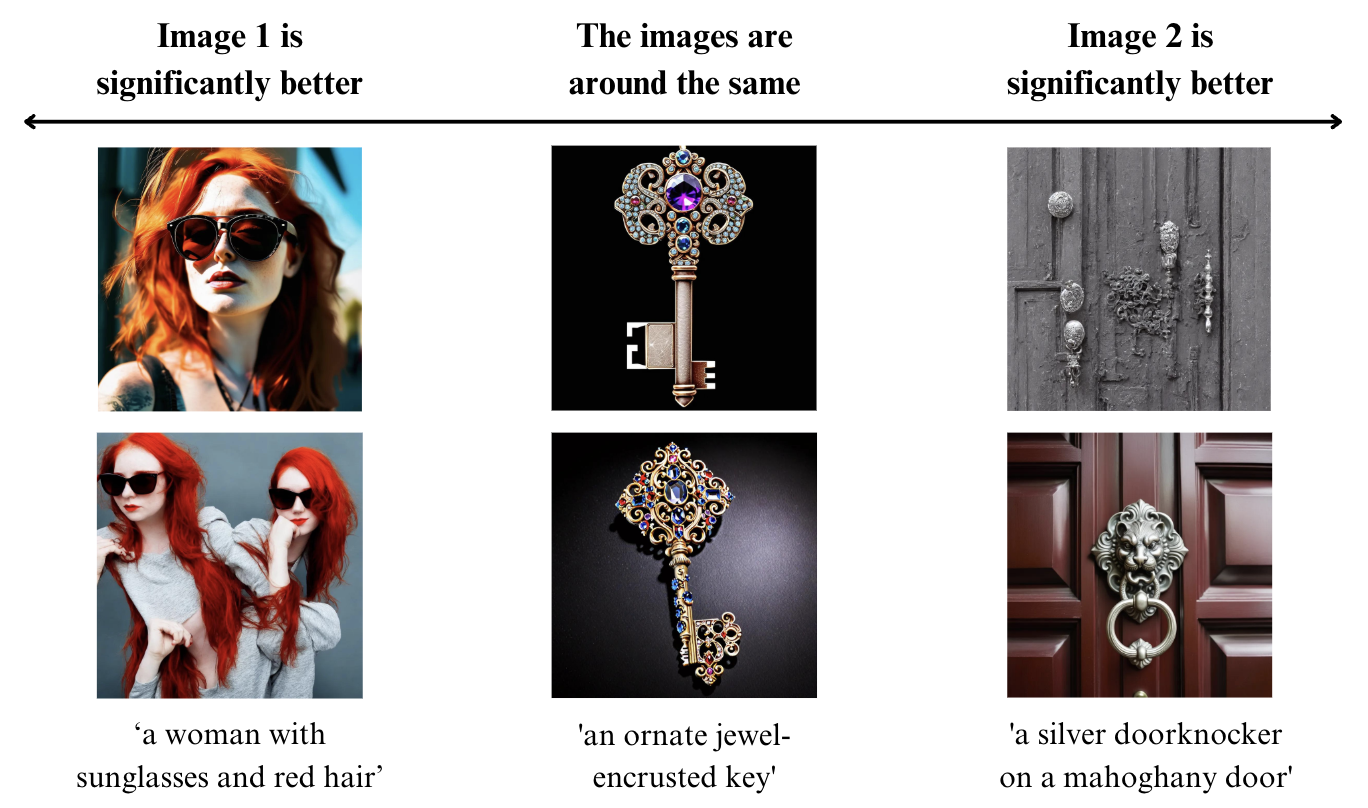}
    \caption{A sample of image pairs across the spectrum of the dataset from similar image quality to each of image 1 and image 2 being significantly better.}
    \label{fig:2}
\end{figure}

\subsection{Verifying Dataset Quality}
\label{sub:verify-noniid-dataset}
After data collection, we verify that the generated dataset has both significant inter-rater alignment and aligns with existing larger human datasets. Since each pair of images is only evaluated by one rater, we evaluate inter-rater alignment using Krippendorff's Alpha (Kripp's Alpha), which is a statistical value measuring whether ratings agree more or less than random~\cite{nassar2021krippsalpha}. Kripp's Alpha is calculated as
\begin{equation}
    \alpha = \frac{p_a - p_e}{1 - p_e}
\end{equation}
where $p_a$ is the observed probability of agreement and $p_e$ is the expected probability of agreement. We calculate Kripp's alpha over the ratings for each model rather than each image pair as a proxy for evaluating inter-rater agreement.
\begin{figure}[t!]
    \centering
    \includegraphics[width=0.8\linewidth]{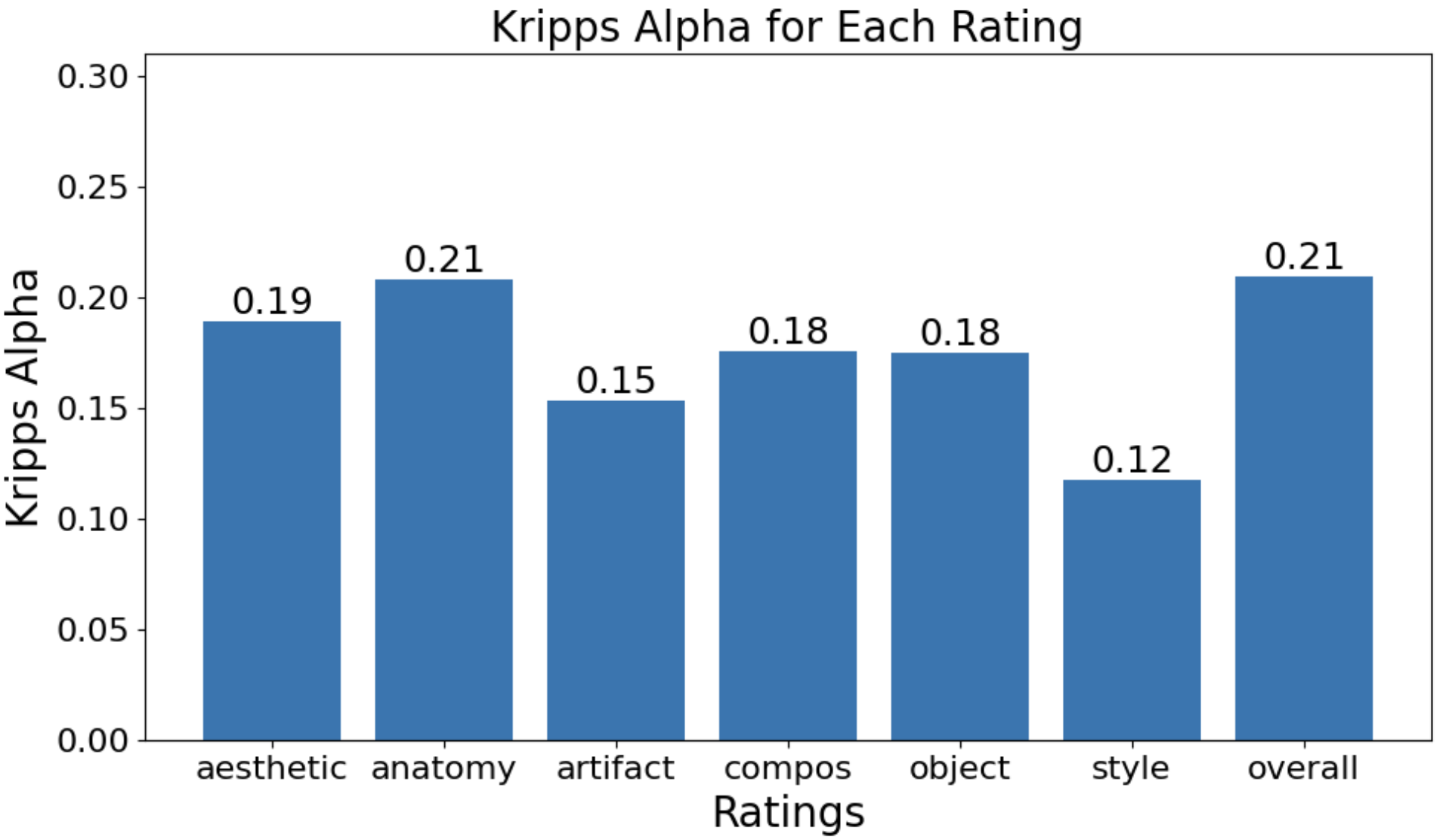}
    \caption{Kripp's Alpha for all seven of the ratings axes evaluated in this study. We find that all of the ratings have Kripp's Alpha significantly above 0, indicating that there is statistically significant inter-rater agreement.}
    \label{fig:3}
\end{figure}

A Kripp's alpha value of 0 indicates negligible agreement, whereas ratings of 1 and -1 indicate perfect agreement and disagreement, respectively. As seen in Fig. \ref{fig:3}, each evaluation axis has Kripp's alpha far above 0, indicating good inter-rater agreement. We note that style and artifacts have lower Kripp's alpha than the other categories -- we attribute this to raters frequently saying that these categories were not relevant, reducing the number of samples and increasing noise. Specifically, when only certain prompts explicitly denoted a style, such as ``in the style of Picasso'', or when neither image had diffusion generated artifacts, users frequently selected ``Not Relevant''. 

\begin{figure}[t!]
    \centering
    \includegraphics[width=0.9\linewidth]{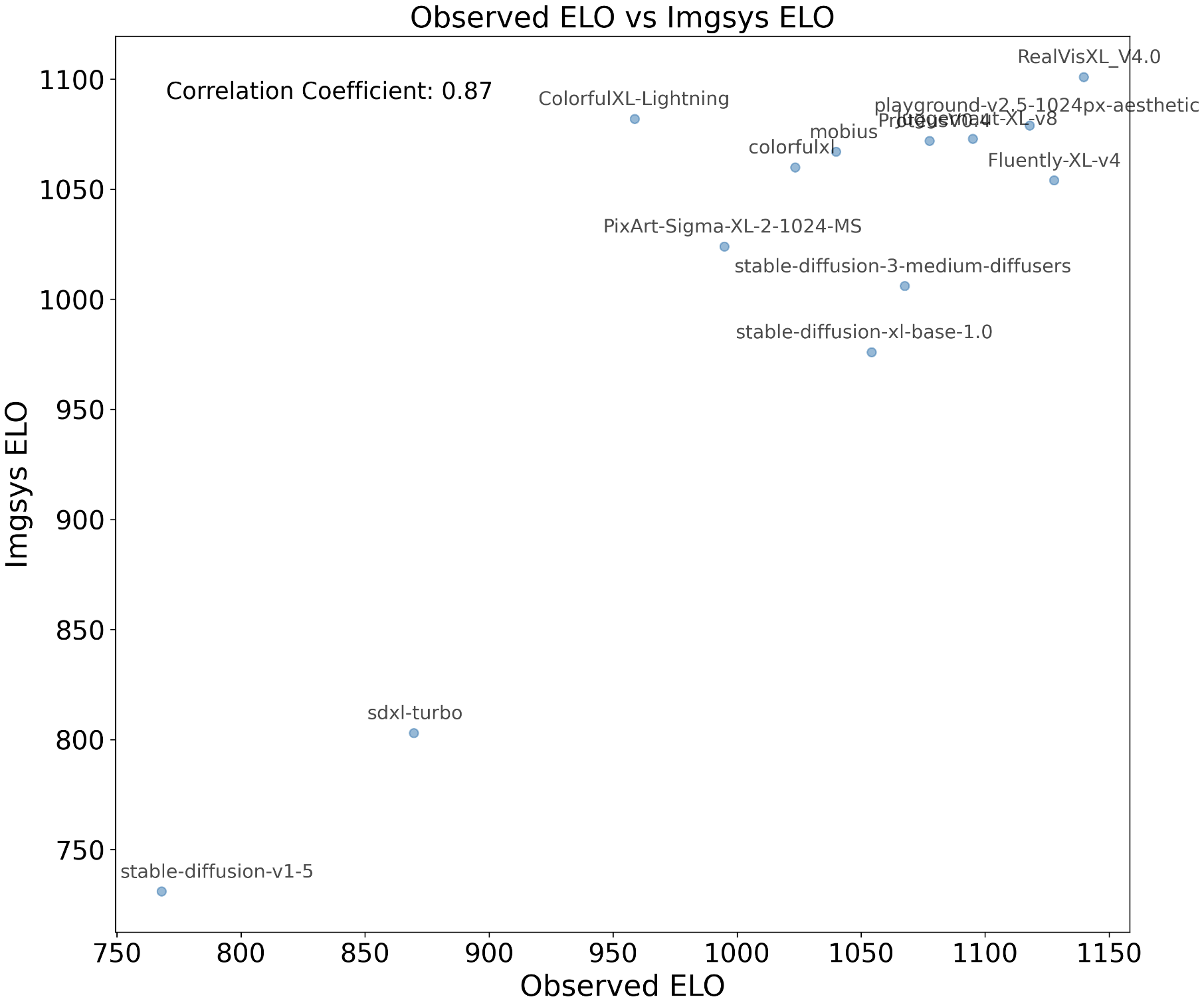}
    \caption{We find that ImgSys Elo scores are strongly correlated with Elo scores calculated from dataset.}
    \label{fig:4}
\end{figure}

We also compare model Elo on our dataset against the larger ImgSys dataset in Fig~\ref{fig:4}, where ImgSys is a standard arena for rating generative image models~\cite{zheng2023llmasajudge, imgsys2025}. This serves as a sanity check to validate our rater pool and data collection methodology. We expect that the ELO rankings in the two datasets should be correlated if our raters are similar to the much larger ImgSys pool. Indeed, we find that ImgSys Elos are highly correlated (Pearson's correlation coefficient $=0.87$) with the overall Elos calculated on our multi-image dataset, suggesting our pool of raters has similar preferences to ImgSys raters. Finally, we quantify potential bias in our rating scheme (due to the rater pool consisting only of AI researchers) by taking a subset of the dataset and comparing ratings with a dataset of non-AI researcher ratings (college students). We find that the correlation ranges from 0.75 and 0.88 across categories and Cohen's Kappa ranges from 0.34 to 0.7, indicating significant agreement.

\renewcommand{\arraystretch}{0.8}

\section{Results on Frontier Models}
\label{multi-task-eval}

In this section, we aim to answer these two guiding questions: (1) how do large frontier models perform on an ensemble of image evaluation tasks, (2) what kind of relationships do large frontier models discover between image quality attributes and how do those compare to humans? We also compare to several smaller models finetuned image reward models on the overall rating task, but these models are unable to provide finer grained results.

\subsection{Prompting Strategy}
LLMs and specialty image reward models require different prompting strategies due to their different training and architectures. Large frontier models can accept multiple images as inputs, so we prompt these models with the two images and request them to output results on a five-item Likert scale specified using special tokens (identical to the dataset collection). These special tokens can then easily be parsed into an identical Likert scale as that used by the dataset.

For single-image finetuned reward models, we instead pass each image into the model separately. The image with the higher rating is considered to be better. We detail the specific prompts used in Appendix Section 5.

\subsection{Frontier Model Evaluation}

We evaluate models using two metrics: the Pearson correlation coefficient (Tab.~\ref{tab:pearson-correlations}) and Cohen's Kappa (Tab.~\ref{tab:kappa-agreement}). Cohen's Kappa reflects rater agreement, whereas Pearson correlation describes the similarity of the two distributions. 

\begin{table*}[t!]
\centering
\begin{tabular}{lccccccc}
\toprule
Model & Aesthetic & Composition & Style & Artifacts & Anatomy & Objects & Overall \\
\midrule
gpt4o & \textbf{0.127} & \textbf{0.147} & \textbf{0.110} & \textbf{0.117} & \textbf{0.152} & \textbf{0.229} & \textbf{0.176} \\
pickscore & - & - & - & - & - & - & 0.164 \\
claude & 0.119 & 0.114 & 0.095 & 0.104 & 0.147 & 0.207 & 0.140 \\
hpsv2 & - & - & - & - & - & - & 0.133 \\
imagereward & - & - & - & - & - & - & 0.128 \\
vqascore & - & - & - & - & - & - & 0.110 \\
gemini-1.5-flash & 0.051 & 0.047 & 0.101 & 0.062 & 0.075 & 0.138 & 0.071 \\
gpt4o-mini & 0.042 & 0.091 & 0.055 & 0.056 & 0.046 & 0.170 & 0.063 \\
pixtral & 0.030 & 0.023 & 0.028 & 0.013 & 0.017 & 0.041 & 0.045 \\
\bottomrule
\end{tabular}
\caption{Cohen's Kappa agreement between human and LLM ratings across different aspects (sorted by Overall).}
\label{tab:kappa-agreement}
\end{table*}

\begin{table*}[t!]
\centering
\begin{tabular}{lccccccc}
\toprule
Model & Aesthetic & Composition & Style & Artifacts & Anatomy & Objects & Overall \\
\midrule
pickscore & - & - & - & - & - & - & \textbf{0.498} \\
gpt4o & \textbf{0.376} & \textbf{0.467} & \textbf{0.337} & 0.349 & \textbf{0.398} & \textbf{0.501} & 0.461 \\
claude & 0.338 & 0.397 & 0.336 & \textbf{0.361} & 0.390 & 0.467 & 0.407 \\
hpsv2 & - & - & - & - & - & - & 0.398 \\
imagereward & - & - & - & - & - & - & 0.398 \\
vqascore & - & - & - & - & - & - & 0.356 \\
gemini-1.5-flash & 0.260 & 0.339 & 0.282 & 0.307 & 0.337 & 0.360 & 0.362 \\
gpt4o-mini & 0.211 & 0.259 & 0.152 & 0.175 & 0.167 & 0.350 & 0.286 \\
pixtral & 0.052 & 0.119 & 0.080 & 0.053 & 0.074 & 0.154 & 0.140 \\
\bottomrule
\end{tabular}
\caption{Pearson correlation between human and LLM ratings across different aspects (sorted by Overall).}
\label{tab:pearson-correlations}
\end{table*}

We find that across all tasks, GPT-4o~\cite{gpt4o} outperforms the other foundation models. However, the specially trained PickScore~\cite{kirstain2023pick}, HPSv2~\cite{wu2023human}, VQAScore~\cite{lin2024vqascore}, and ImageReward~\cite{xu2023imagereward} outperform Gemini-1.5-Flash~\cite{gemini15}, GPT-4o-mini~\cite{gpt4o-mini}, and Pixtral~\cite{agrawal2024pixtral} on the overall task, showing that fine-tuned reward models can still outperform foundation models given enough human preference data. However, unlike foundation models, these small models cannot perform tasks outside of overall judgement, as these are out of the models' domain. 

We also find that object adherence is the easiest of the six subtasks, as the Pearson correlation and Cohen's Kappa are the highest. We also find that style, artifacts, and aesthetic are consistently more difficult judgment subtasks.

\subsection{Correlating Image Quality Attributes}
\label{subsec:attribute-correlations}
To evaluate correlations between image quality attributes, we calculate the Pearson correlation coefficient between tasks for both human ratings and the best performing general LLMs: GPT-4o and Claude-3.5-Sonnet.

\begin{figure}[h!]
    \centering
    \includegraphics[width=0.6\linewidth]{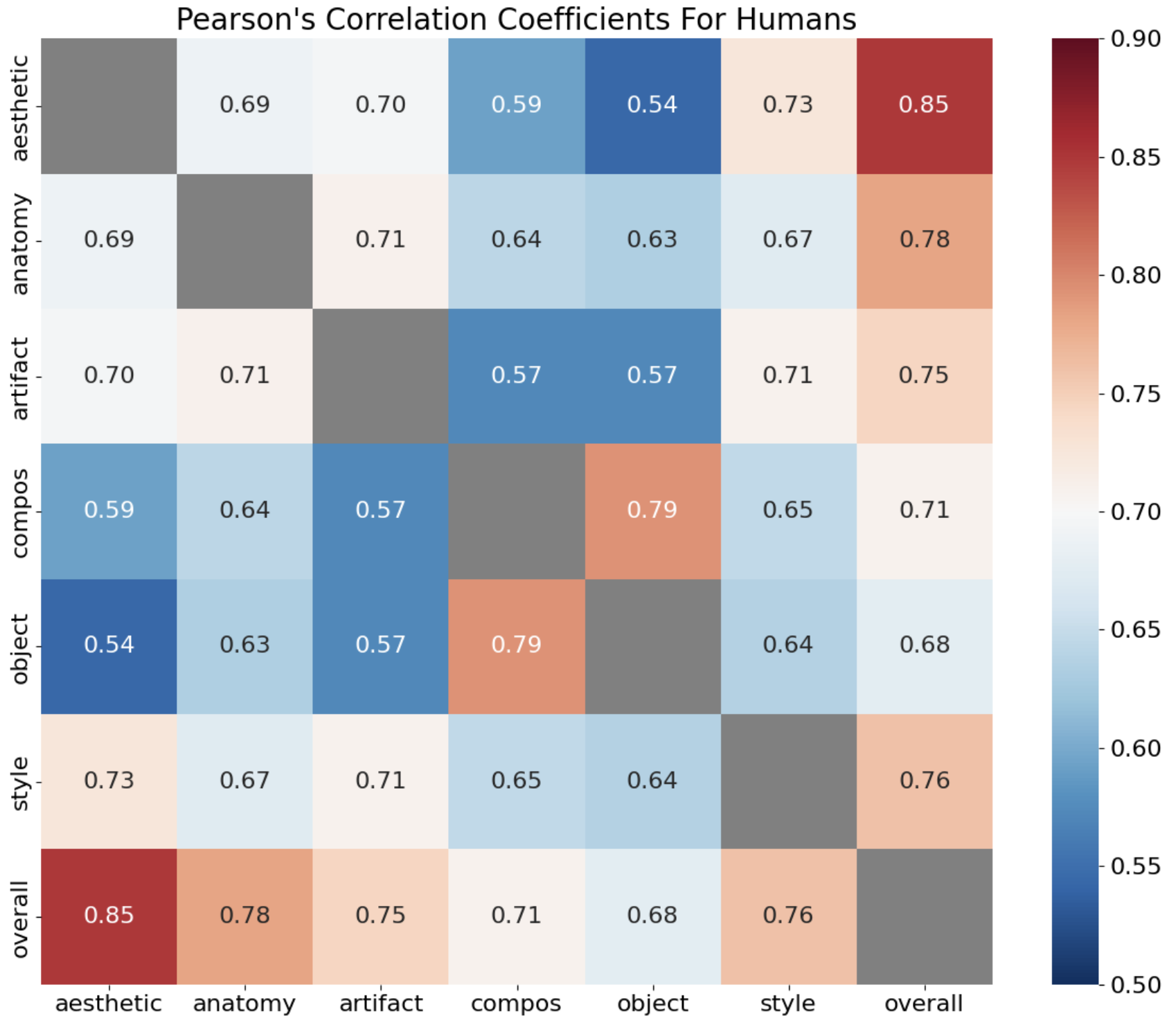}
    \caption{Correlation coefficients for human ratings, demonstrating various strong linkages between pairs of tasks.}
    \label{fig:human-pearsons}
\end{figure}

As shown in Fig. \ref{fig:human-pearsons}, humans exhibit a strong correlation between overall ratings and all attributes other than object adherence. This possibly signifies that object adherence is relatively unimportant for human image preferences, which is interesting given that object adherence is the easiest task for LLMs to evaluate. On the other hand, aesthetics are highly correlated with overall rating. We also find correlations between individual tasks. For example, human raters strongly associate compositional correctness with object adherence, and we observe weak associations between aesthetic and style, as well as between anatomical issues and diffusion artifacts.


\begin{figure}[h!]
    \centering
    \includegraphics[width=0.6\linewidth]{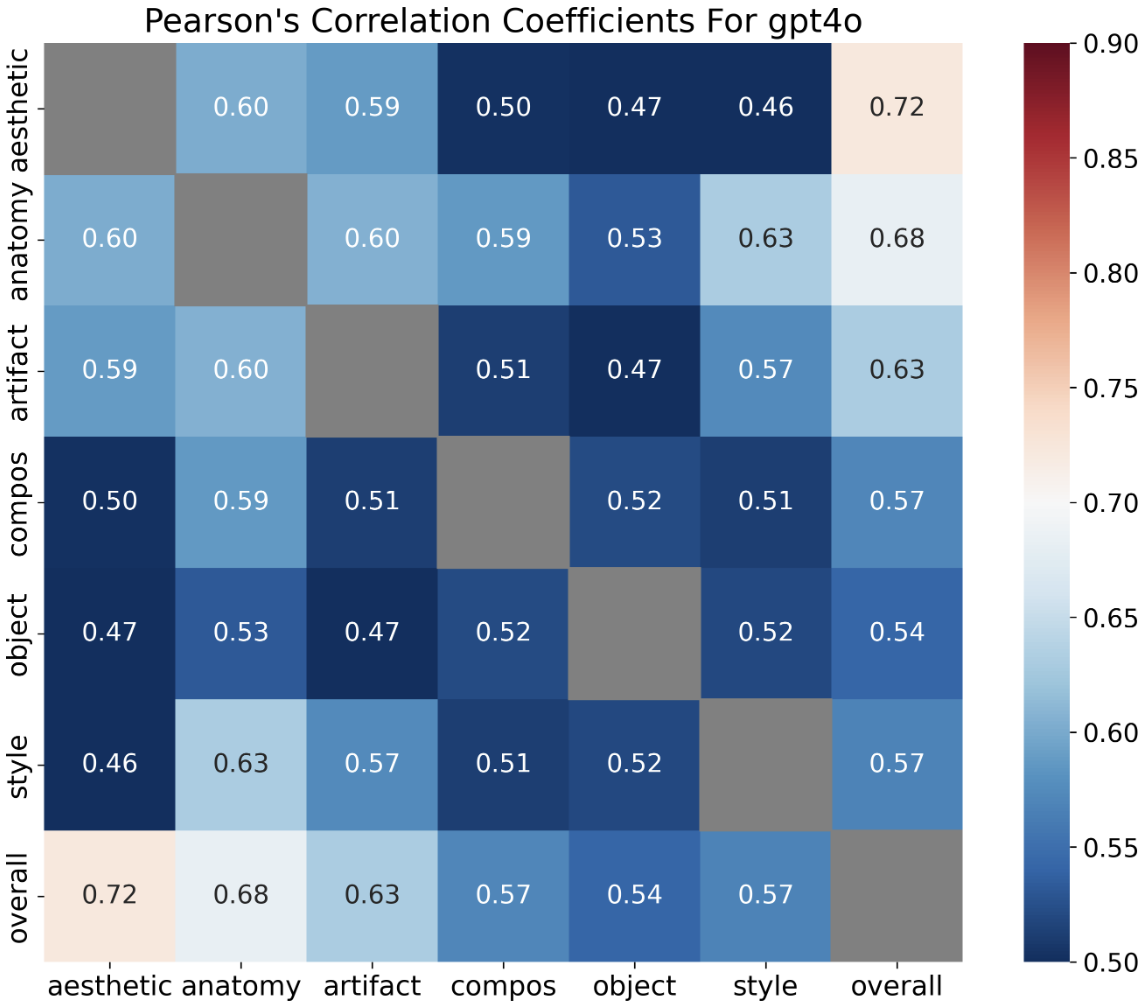}
    \caption{Correlation coefficients for GPT-4o ratings, demonstrating a lack of strong link between tasks.}
    \label{fig:gpt4o-pearsons}
\end{figure}

\begin{figure}[h!]
    \centering
    \includegraphics[width=0.6\linewidth]{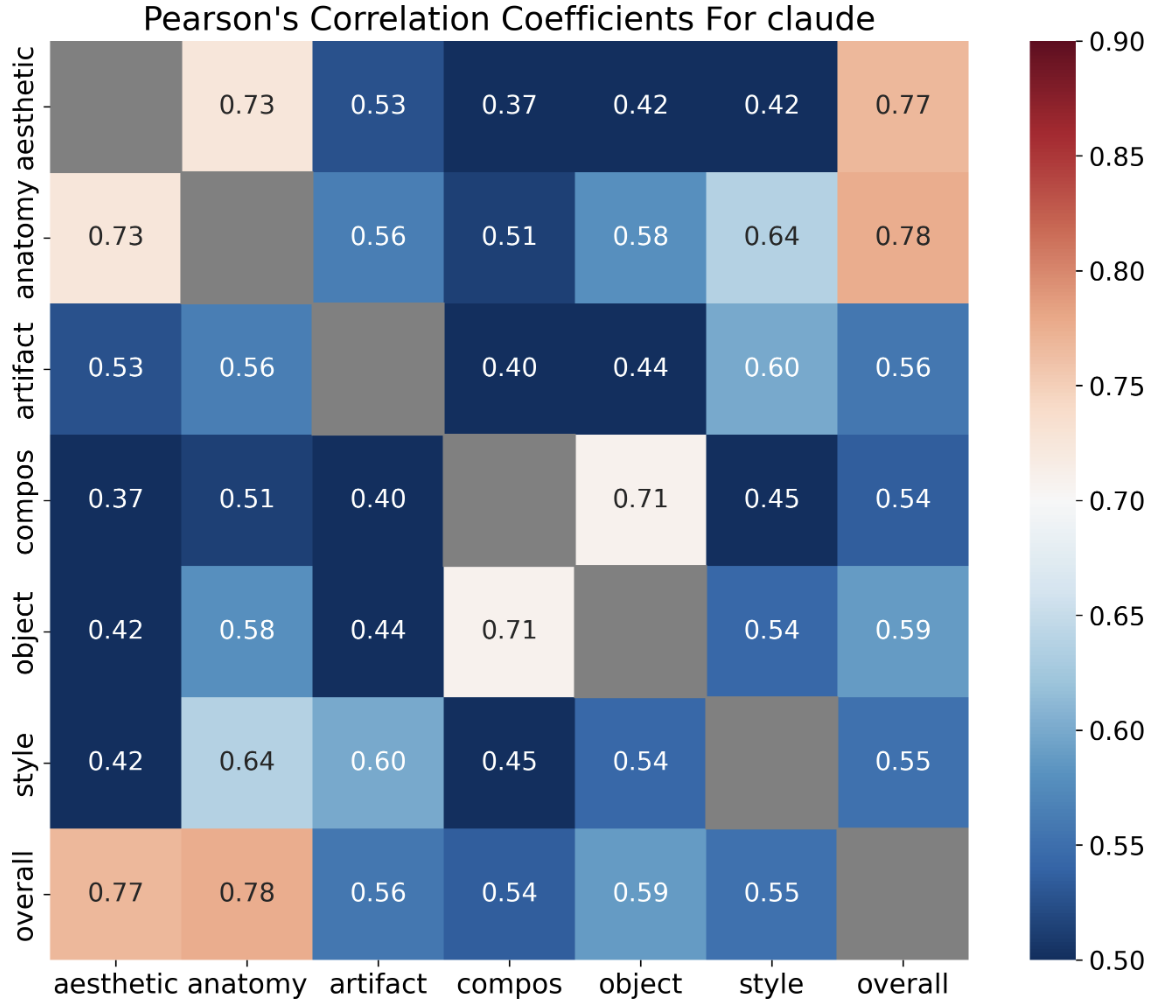}
    \caption{Correlation coefficients for Claude ratings, demonstrating a strong link between aesthetic and anatomy, and compositionality and object adherence.}
    \label{fig:claude-pearsons}
\end{figure}

Interestingly, LLMs lack a similar correlation structure. The preference for aesthetic is still present in both GPT-4o and Claude-3.5-Sonnet (Claude) as exhibited by Fig. \ref{fig:gpt4o-pearsons} and Fig. \ref{fig:claude-pearsons}, but Claude shows a  high correlation between object adherence and the overall rating, in contrast to humans, who show a higher overall correlation for style or artifacts. On top of that, GPT-4o seems to have extremely weak correlations between each pair of non-overall tasks, which is in stark contrast to the human case. These differences highlight that while humans rate images based on complex image quality attribute relationships, LLMs have not yet achieved similar ways of reasoning about these attributes.

\section{Synthetic Attribute Datasets}
\label{sec:synth-datasets}
\begin{figure}[t!]
    \centering
    \includegraphics[width=\linewidth]{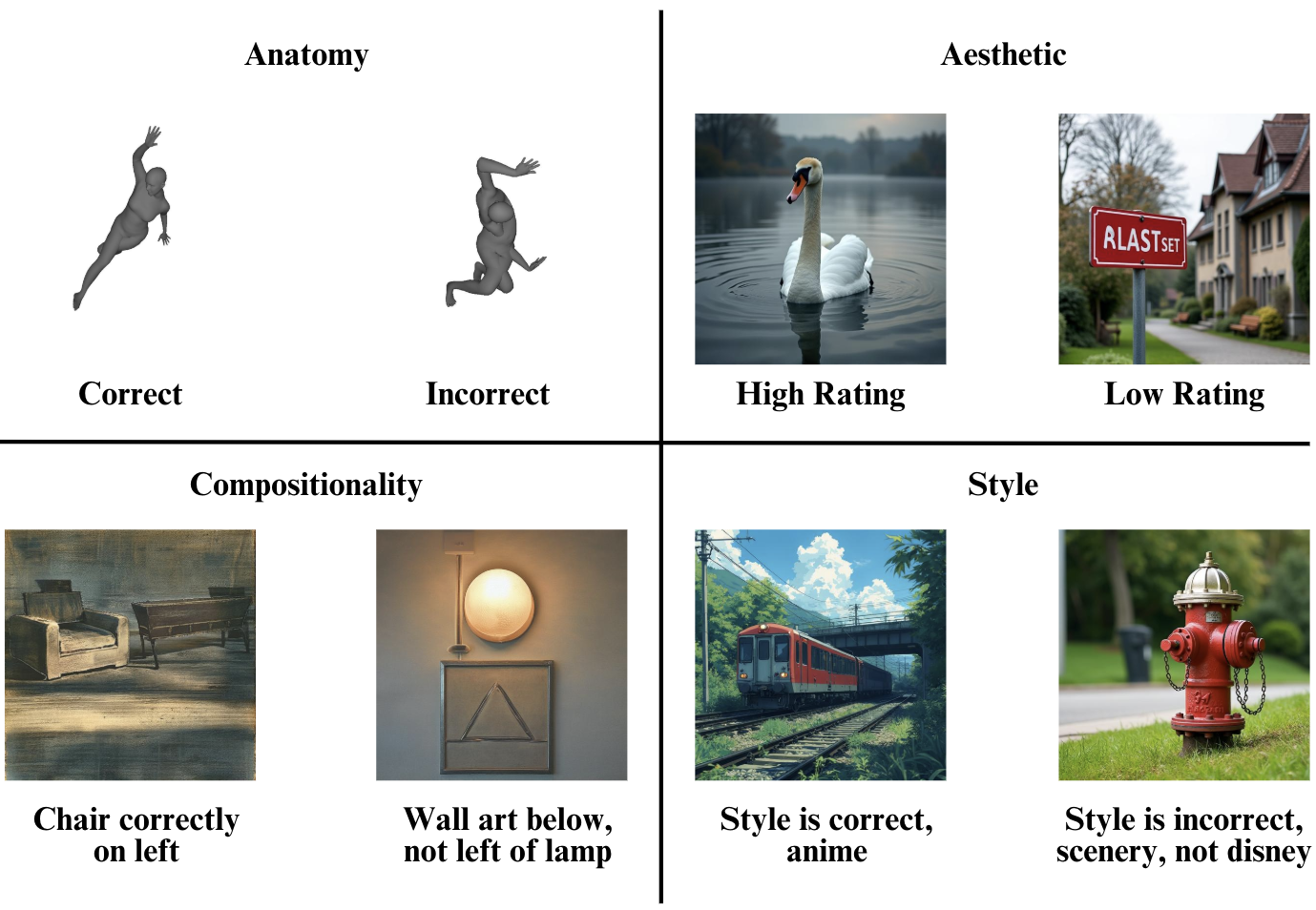}
    \caption{Sample correct and incorrect images from each of the synthetic datasets.}
    \label{fig:synthetic-data}
\end{figure}
As we have seen in Section \ref{subsec:attribute-correlations}, LLMs struggle to understand the relationship humans display between image quality attributes and overall image quality. This raises questions about the degree to which LLMs understand each individual image quality attribute in isolation. To explore this, we construct separate subtasks for four of the image quality attributes previously discussed: aesthetic, composition, style, and anatomy. For these attributes it is possible to generate large amounts of synthetic data, so we can build training and evaluation datasets for each subtask with high quality labels. Each of these simple subtasks is intended to be a task where humans can easily perform well (e.g. identifying correct and incorrect human geometry) but LLMs may struggle. We individually describe each of the datasets below and provide examples of sample data in Fig. \ref{fig:synthetic-data}. Further examples are presented in Appendix Section 4.

\subsection{Synthetic Aesthetic Dataset}
\label{subsec:synth-aes}
\textbf{Purpose:} To study whether LLMs can easily model the judgement of aesthetic quality, using the AesExpert model as a proxy~\cite{huang2024aesexpert}.
\newline\textbf{Data Generation:} We sample 50K prompts from COCO prompts and generate images using FLUX-1.0[dev]~\cite{lin2014coco, flux1dev2024}. Each of these images is rated by the AesExpert model~\cite{huang2024aesexpert}.
\newline\textbf{Prompt Format:} We provide the model with the image and original image prompt, and ask the model to output a floating point rating between 1 and 10.

\subsection{Synthetic Anatomy Dataset}
\label{subsec:synth-anat}
\textbf{Purpose:} To study whether LLMs can detect simple distortions in human anatomy.
\newline\textbf{Data Generation:} We use the parameterized articulated SKEL 3D model of a human body~\cite{keller2023skel}. By varying joint parameters, 50k samples are generated in a 50/50 split where half of the samples match normal human joint parameters, while half of the samples do not. These samples are rendered using OpenGL with a fixed camera.
\newline\textbf{Prompt Format:} This task provides the model with an image of a generated human body, and asks the model whether the human is anatomically correct or distorted.

\subsection{Synthetic Composition Dataset}
\label{subsec:synth-comp}
\textbf{Purpose:} To study whether LLMs can identify the four cardinal directions, a fundamental compositional task.
\newline\textbf{Data Generation:} We sample 50k image pairs of household objects taken from the Amazon Object (ABO) dataset~\cite{collins2022abo}. These pairs are converted into Canny edge maps, which are combined based on the cardinal direction of choice. We convert the combined Canny edges into images using a FLUX-1.0[dev] ControlNet~\cite{flux1dev2024, zhang2024controlnet}. The image is screened using Florence-2 to evaluate whether the original objects taken from the ABO dataset match those generated in the image, ensuring output quality~\cite{xiao2024florence2}.
\newline\textbf{Prompt Format:} This task provides the model with the image and a prompt which may or may not have the correct direction, and asks the model if the direction is properly displayed in the image.

\subsection{Synthetic Style Dataset}
\label{subsec:synth-style}
\textbf{Purpose:} With this dataset, we evaluate whether LLMs can distinguish six styles that are distinguishable for humans.
\newline\textbf{Data Generation:} We sample 10k images each from these six styles: anime, art, disney, midjourney, realism, and scenery. Images are generated using prompts from COCO prompts and a LoRA version of FLUX-1.0[dev]~\cite{lin2014coco, flux1dev2024, hu2022lora}.
\newline\textbf{Prompt Format:} This task provides the model with a prompt that may or may not have the correct style, asking the model whether or not the style direction is properly displayed in the image.
\begin{table*}
\centering
\begin{tabular}{lrrrr}
\toprule
Model & Anatomy Accuracy & Compo Accuracy & Style Accuracy & Aesthetic Correlation \\
\midrule
\textbf{Human Evaluator} & \textbf{78.2\%} & \textbf{99\%} & \textbf{79\%} & \textbf{0.671} \\
\textbf{Best Finetuned Model} & \textbf{57.6\%} & \textbf{91.2\%} & \textbf{81.8\%} & \textbf{0.847} \\
claude-3-5-sonnet-20241022 & \underline{52.0\%} & 71.0\% & 60.4\% & 0.257 \\
gpt-4o & 51.6\% & \underline{75.2\%} & \underline{61.8\%} & \underline{0.583} \\
gpt-4o-mini & 47.7\% & 58.6\% & 61.0\% & 0.586 \\
gemini-1.5-flash-002 & 47.8\% & 57.4\% & 62.7\% & 0.037 \\
pixtral-12b & 51.2\% & 46.6\% & 55.7\% & -0.053 \\
baseline LLaVA & 50.0\% & 52.0\% & 57.2\% & -0.010 \\
\bottomrule
\end{tabular}
\caption{Merged accuracy results across all datasets, demonstrating that anatomy is the most challenging, while compositionality can be easily learned, despite a somewhat lack of capability in large language models.}
\label{tab:merged-results}
\end{table*}
\subsection{Validating Synthetic Data Quality}
\label{subsec-synth-quality}
We perform an independent human evaluation on a random subset of the datasets to validate the quality of our synthetic data. For each of the datasets, we align based on the metric used to evaluate the models, which is accuracy for the Yes/No tasks of anatomy, compositionality, and style, and Pearson correlation for the aesthetic task. For each of the tasks, as presented in Table \ref{tab:merged-results}, the human evaluator does significantly better than random guessing, indicating that the synthetic data has a strong signal for each the task.

\section{Shared Training and Evaluation Details}
\label{sec:training}
For training, we use LLaVA (LLaMA-3-8B LM and 384x384 shape-optimized SigLIP VLM) trained with a 1e-5 learning rate and Adam~\cite{liu2023visual, kingma2015adam, zhai2023siglip}. Models were trained on A100-80GB for around 160 A100-hours each on Transformers 4.44. 

To teach LLaVA multi-image understanding, we use a custom data mix, mixing the multi-image Commonalities and Differences (CaD) dataset with baseline LLaVA data~\cite{lin2024comparison}. We use this data in a third phase of LLaVA training, where the LLM and vision projector are both unfrozen. On top of CaD, we also mix in the training split of our synthetically generated datasets as a small proportion of the training dataset. All models are trained for one epoch, and we sweep across 200k, 400k, and 800k as the number of multi-image samples seen.

We find that including the CaD and LLaVA data creates a push-and-pull effect, where CaD data helps the model acquire multi-image capabilities and LLaVA data prevents the model from losing baseline VQA capabilities. Without LLaVA data, the model catastrophically overfits on either CaD data or the synthetic data provided, even when the synthetic data is at most 10\% of the data mix. In-depth ablations are provided in Appendix Section 6. 

\section{Results on Individual Tasks}
\label{subtask-results}

\subsection{Individual Model Performance}
On these tasks, we evaluate the five large LLMs evaluated in the previous section (GPT-4o, GPT-4o-mini, Claude-3.5-Sonnet, Gemini-1.5-Flash, and Pixtral-12B). As in Table \ref{tab:merged-results} we find that these models perform better than baseline LLaVA, which performs near random on these simple tasks, but do not reach the performance of a human evaluator. 

Pixtral and baseline LLaVA lag behind  most, with performances mostly near random guessing, compared to the strong performance of frontier scale models, indicating that there may be a fundamental change in the performance of multimodal LLMs as image evaluators at some baseline data and parameter scale. However, models specifically finetuned for each task vastly outperform the frontier models, coming near human accuracy on both compositionality and style, and better matching the aesthetic of AesExpert than humans~\cite{huang2024aesexpert}. We find that finetuning on a mix of tasks does not increase performance, with ablations provided in Appendix Section 6.

\subsubsection{Failure Modes on Anatomy}
\begin{figure}
    \centering
    \includegraphics[width=0.6\linewidth]{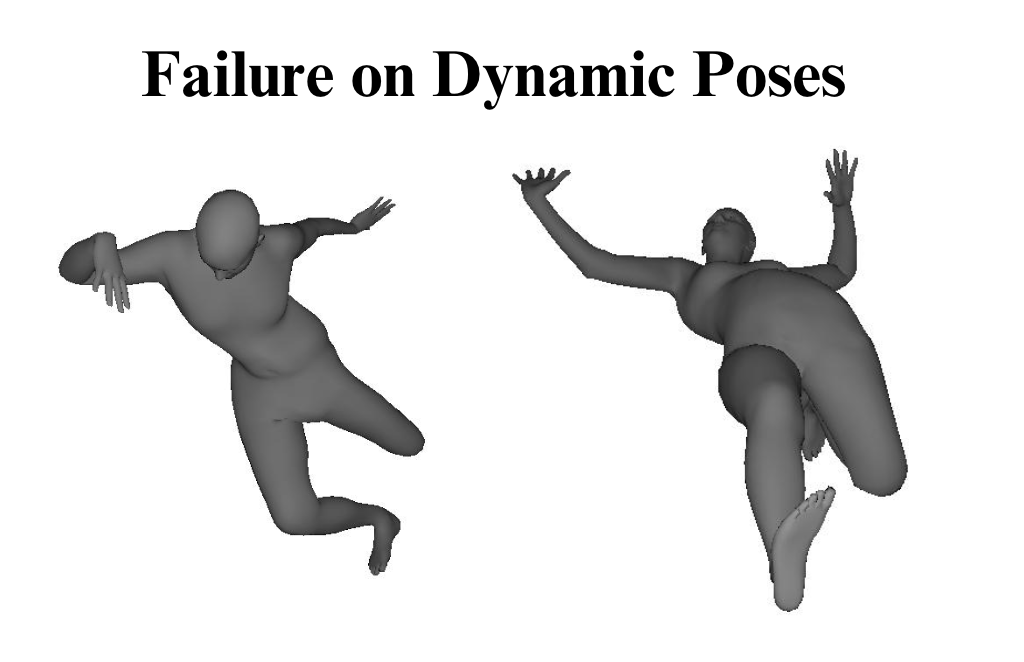}
    \caption{GPT-4o and other LLMs label dynamic poses as anatomically incorrect, an intriguing failure mode.}
    \label{fig:anatomy-failure}
\end{figure}
Even with finetuning, model performance lags behind on anatomy, despite human evaluators performing well. No LLM achieves more than random guessing, which seems to be primarily caused by the failure mode displayed in Fig. \ref{fig:anatomy-failure}. We observe that in anatomically valid dynamic poses, like the two images provided (kicking a soccer ball, and bottom view of running), LLMs mark these poses as anatomically impossible, leading to poor correlation with human raters.

Mechanistically, multimodal LLM (MLLM) performance can be strongly attributed to the alignment of vision-language encoders like CLIP, which are trained on captioning tasks. MLLMs are thus biased to captioning-related tasks like object adherence, and not spatial reasoning. Table 3 shows this behavior, as while anatomy involves the same spatial reasoning as composition (angle and orientation), LLMs consistently underperform, implying inductive biases in the training procedure.

To confirm this, we conduct a further experiment, where MLLMs are prompted to output joint angles before outputting an anatomical decision. Outputted angles are reasonable, but final accuracy remains low, with Gemini and GPT at 46\% and Claude at 54\%. This indicates that while models can reason about angles, they lack inductive bias to correctly process these angles into some more complex result.

\section{Conclusion}
\label{sec:conclusion}
We discover that multimodal LLMs show significant differences from humans in how they rate generated images. Surprisingly, this difference is hidden by their comparable performance at providing overall ratings, only becoming apparent when models and humans are evaluated on individual image quality attributes. Specifically, through our curated multi-task image rating dataset, we find that although large LLMs and smaller reward models are aligned with human preferences, LLMs do not learn the same similarities between pairs of image quality attributes that standard human raters do. Even on simplified versions of image rating tasks, LLMs do not generalize in the same fashion as humans, performing well on tasks like evaluating compositional correctness, but falling far behind in tasks like detecting incorrect anatomy. These findings unveil potential gaps in the overall alignment of multimodal LLMs as image rating judges. The more we rely on automated evaluation for generative image models, the more the need to fill these gaps and ensure that our automated judges are actually aligned with human preferences on an ensemble of tasks.

\bibliography{aaai2026}
\clearpage

\input{suppl_info_section}

\end{document}

%% file: suppl_info_section.tex
\section*{Appendix}
\paragraph{Note:}
We note that all figures and tables are presented after the body of the text due to their size.
\subsection{Human-Rated Dataset Creation and Generation Settings}

For human dataset creation, the models used are\\ \texttt{['mobius'~\cite{corcelio2024mobius}, \\'PixArt-Sigma-XL-2-1024-MS'~\cite{chen2024pixart}, \\'Juggernaut-XL-v8'\\~\cite{rundiffusion2024juggernaut}, \\'RealVisXL\_V4.0'~\cite{sg1612222024realvisxl}, \\'ProteusV0.4'~\cite{dataautogpt32024proteus}, \\ 'Fluently-XL-v4'~\cite{fluently2024fluentlyxl}, \\'playground-v2.5-1024px-aesthetic'~\cite{li2024playgroundv2.5}, \\'ColorfulXL-Lightning'~\cite{recoilme2024colorfulxllightning}, \\'colorfulxl'~\cite{recoilme2024colorfulxl}, \\'stable-diffusion-v1-5'~\cite{runwayml2022stablediffusionv15},\\ 'sdxl-turbo'~\cite{sauer2024sdxlturbo}, \\'stable-diffusion-2-1'~\cite{rombach2021high},\\ 'stable-diffusion-2-base'~\cite{rombach2021high}, \\'stable-diffusion-3-medium-diffusers' ~\cite{esser2024sd3}, \\'stable-diffusion-xl-base-1.0'~\cite{podell2023sdxl}]}. \\Each of these models was run on the entire PartiPrompts dataset~\cite{yu2023parti}, at an image size of 1024x1024, 50 steps and guidance scale of 7.

\begin{figure*}
    \centering
    \includegraphics[width=0.9\linewidth]{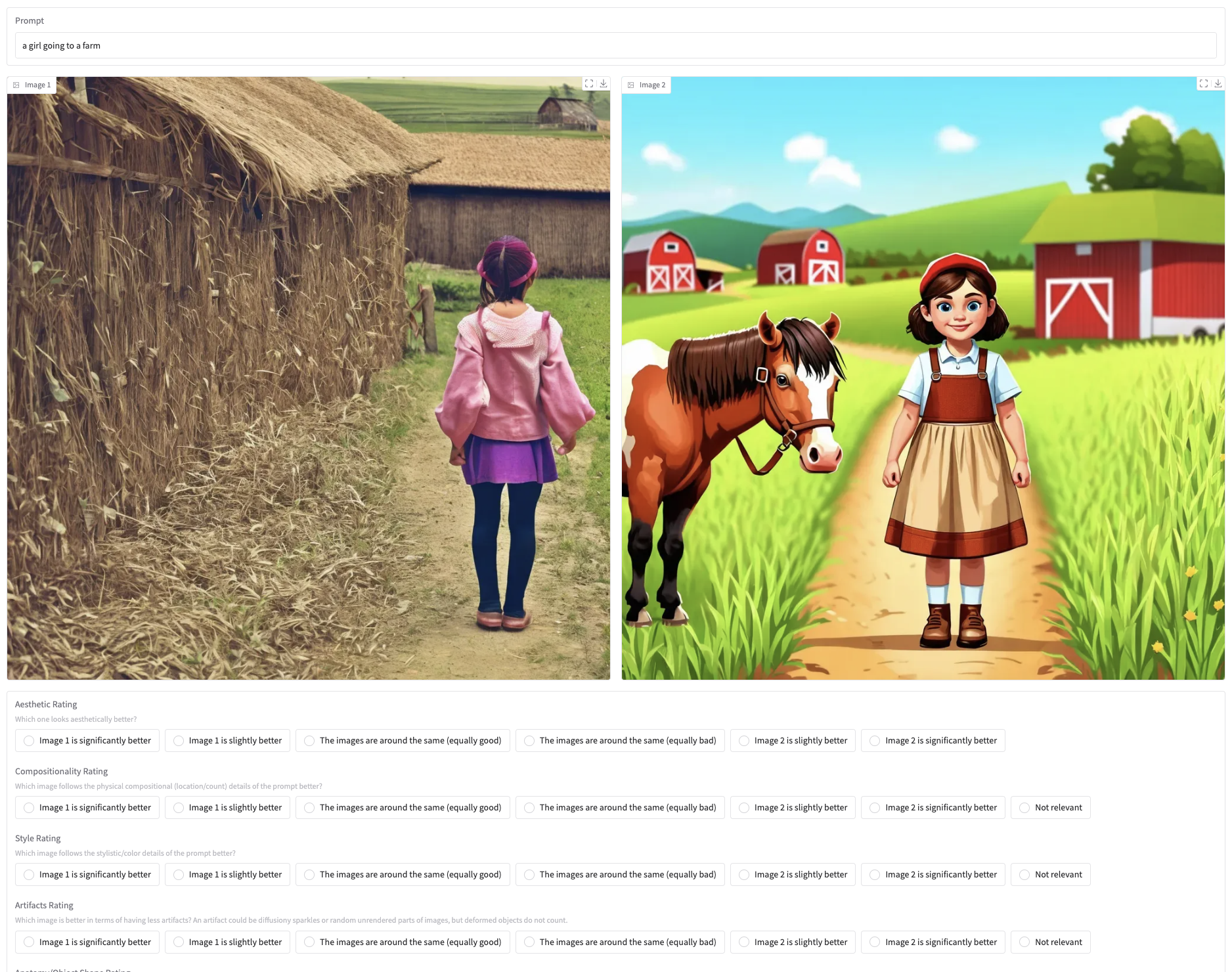}
    \caption{The Gradio UI provided to human raters with images and prompt on top and different rating options below.}
    \label{suppl_fig:gradio-ui}
\end{figure*}

Pairs of these images were then provided to raters via a Gradio survey, which is shown in~\Cref{suppl_fig:gradio-ui}. The raters consist of 40+ AI researchers who each rated on average 30 images.

\subsection{Task Definition and Reduction of Ambiguity}
We provide the specific descriptions used of the tasks here. All were provided to both LLM and human annotators.

\begin{enumerate}
    \item Aesthetic: ``aesthetically better''
    \item Compositionality: ``physical compositional (location/count) details of the prompt''
    \item Style: ``stylistic/color details''
    \item Artifacts: ``diffusiony sparkles or random unrendered parts of images, but deformed objects do not count.''
    \item Anatomy: ``better anatomy/object shape (e.g. no deformation or missing parts)''
    \item Object Adherence: ``relevant objects from the prompt''
\end{enumerate}

We believe that prompts disambiguate the various topics of the tasks, limiting concept confusion of annotators during the rating process. Human annotators were also instructed to list ``Not relevant'' whenever certain tasks were not present (like anatomy), which further serves to reduce concept confusion in the rating process. 
\subsection{Sample Dataset Images}

\begin{figure*}
    \centering
    \includegraphics[width=0.9\linewidth]{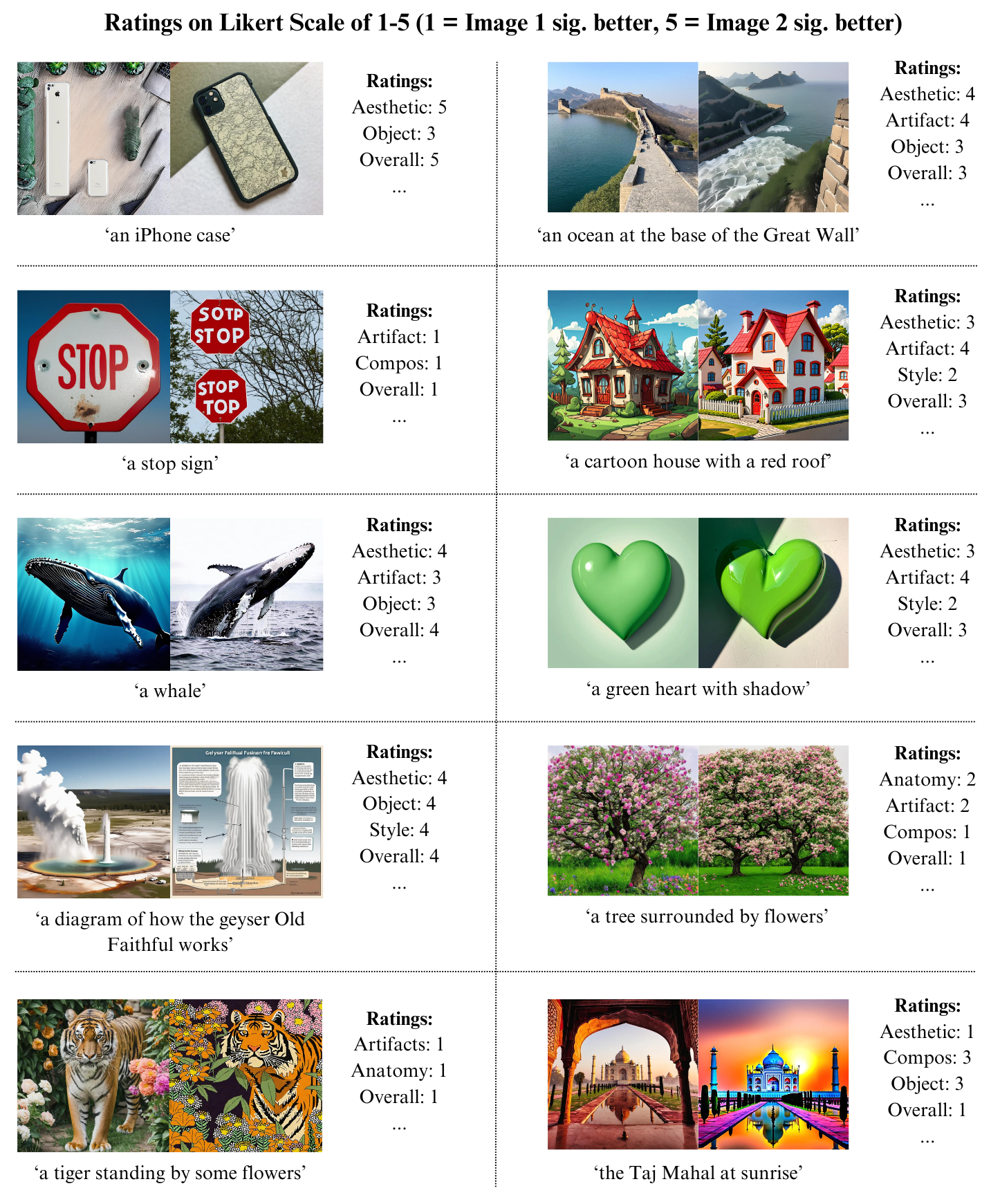}
    \caption{Additional examples of human ratings on pairs of diffusion generated images.}
    \label{suppl_fig:extra-human-ratings}
\end{figure*}

We provide additional examples of annotations on diffusion generated images rated by humans in~\Cref{suppl_fig:extra-human-ratings}.

\subsection{Synthetic Dataset Examples}
\begin{figure*}
    \centering
    \includegraphics[width=0.9\linewidth]{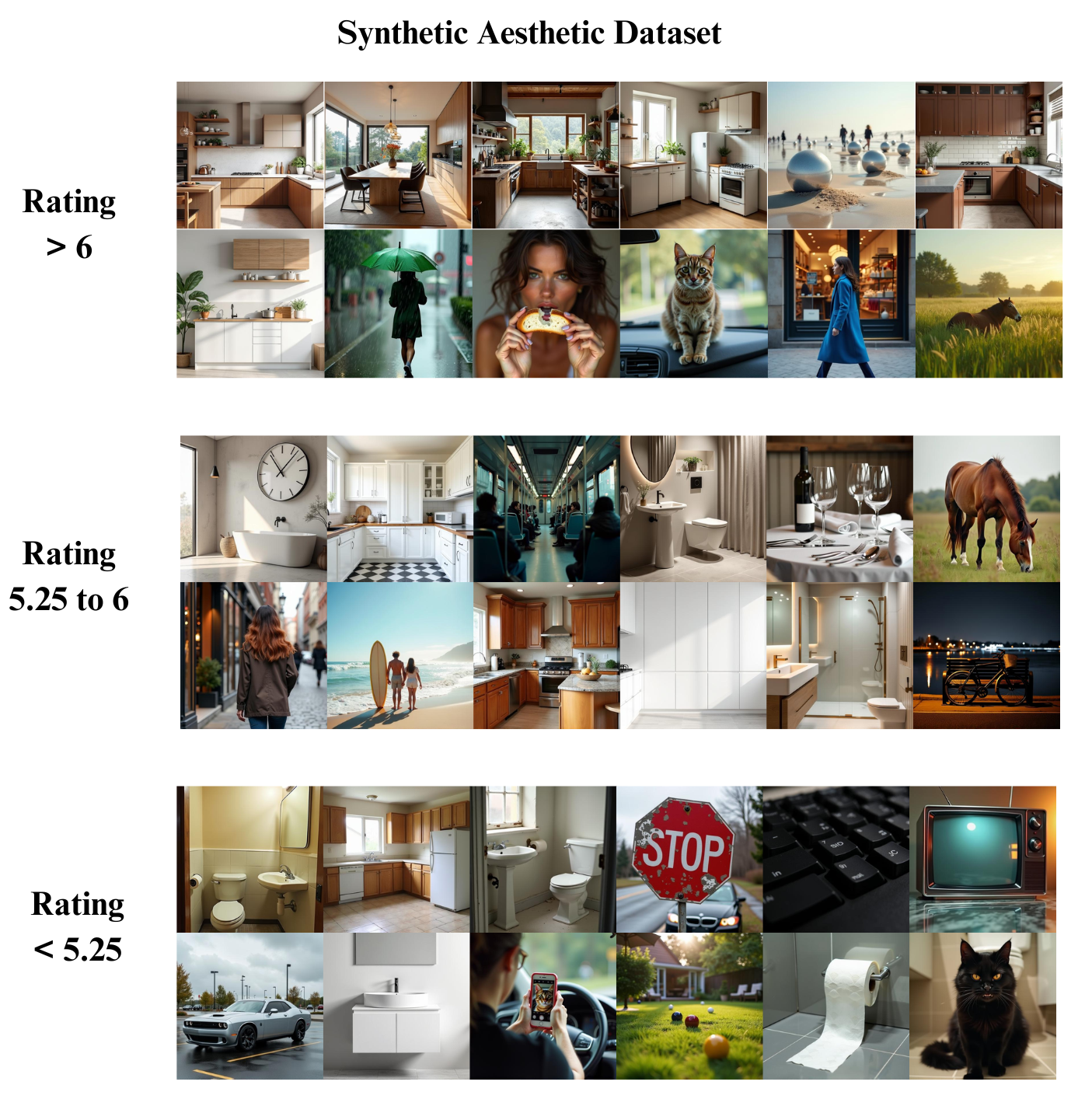}
    \caption{Additional examples of the synthetic aesthetic dataset.}
    \label{suppl_fig:synth-aes}
\end{figure*}

\begin{figure*}
    \centering
    \includegraphics[width=0.9\linewidth]{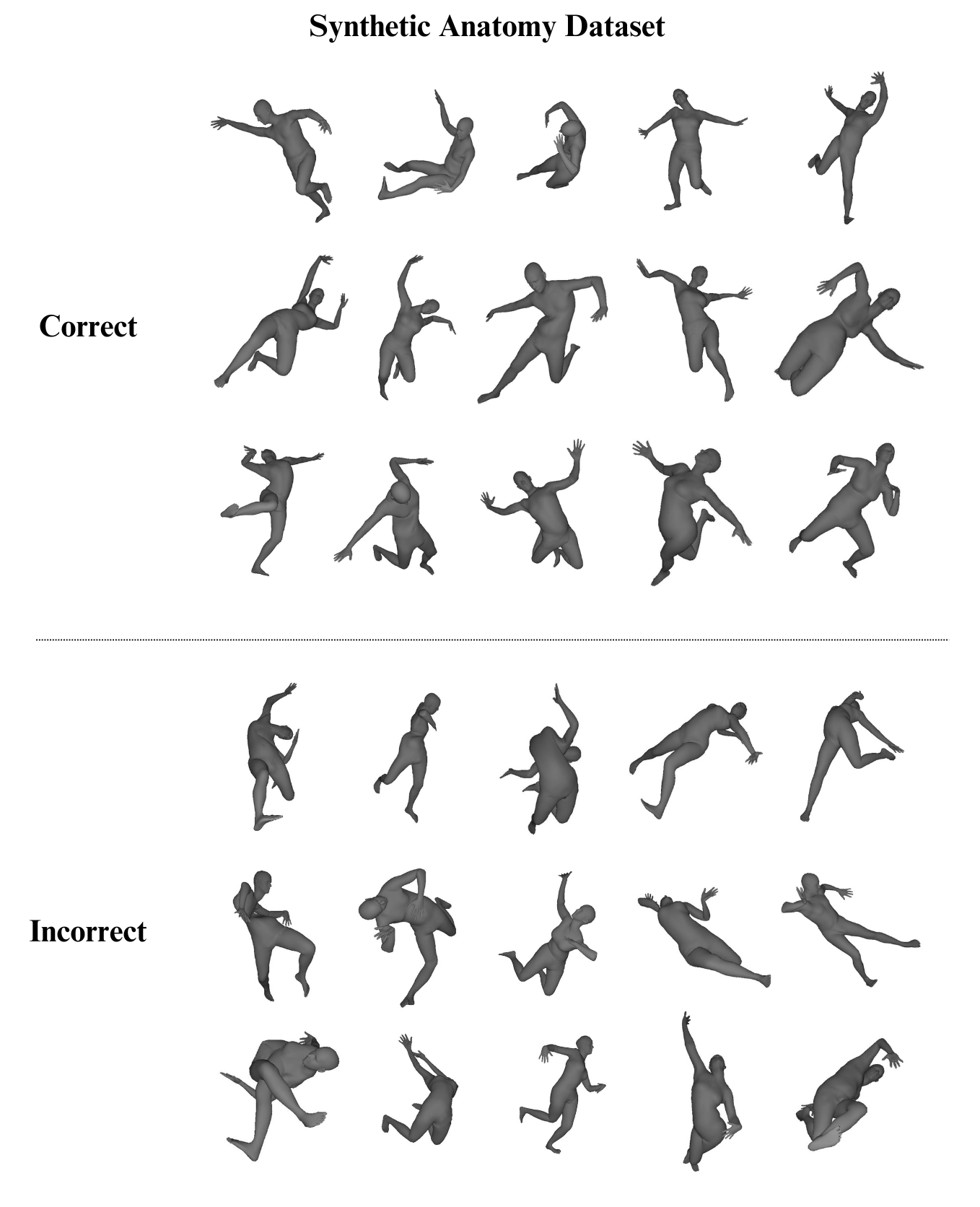}
    \caption{Additional examples of the synthetic anatomy dataset.}
    \label{suppl_fig:synth-anatomy}
\end{figure*}

\begin{figure*}
    \centering
    \includegraphics[width=0.9\linewidth]{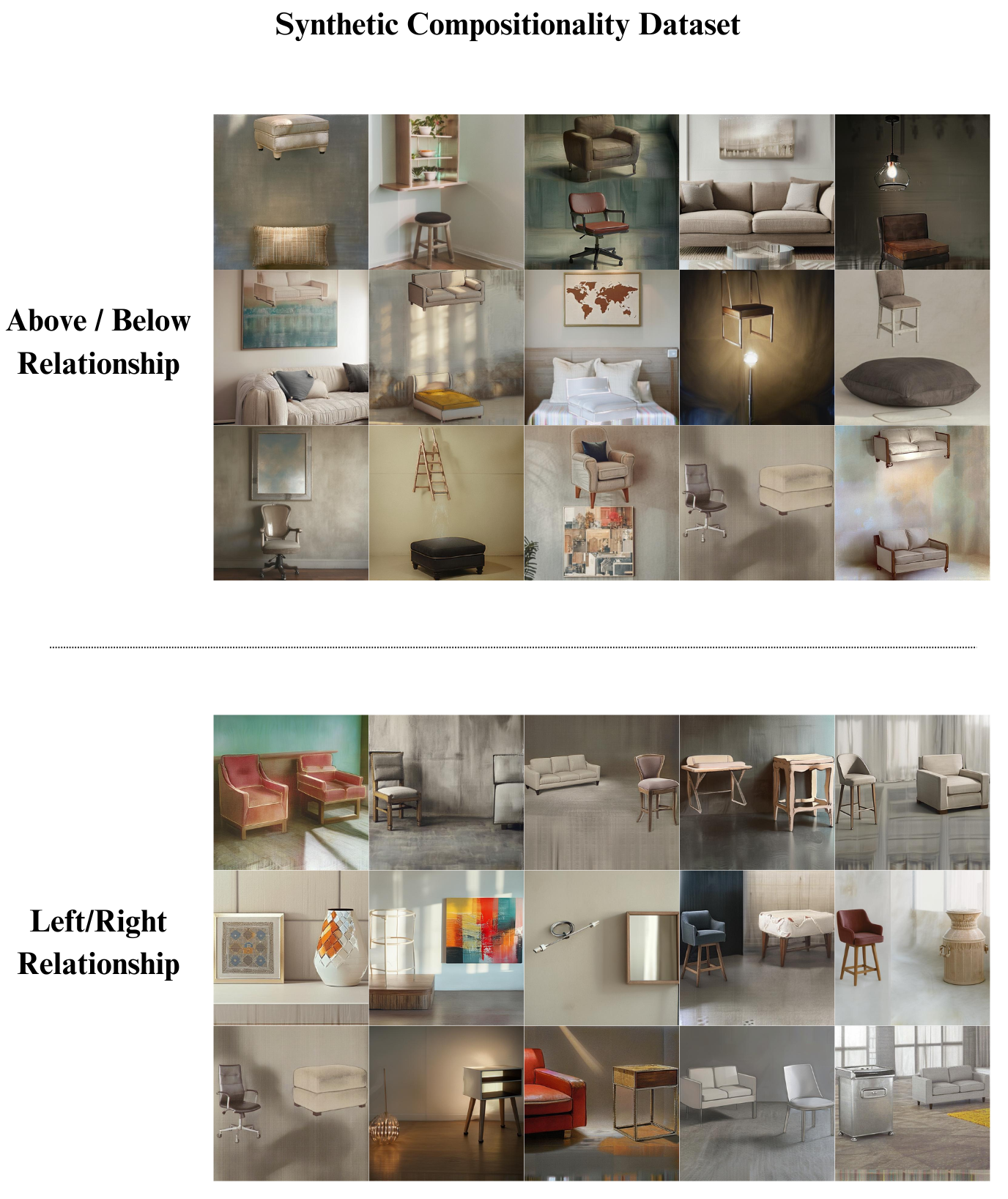}
    \caption{Additional examples of the synthetic compositionality dataset.}
    \label{suppl_fig:synth-compo}
\end{figure*}

\begin{figure*}
    \centering
    \includegraphics[width=0.9\linewidth]{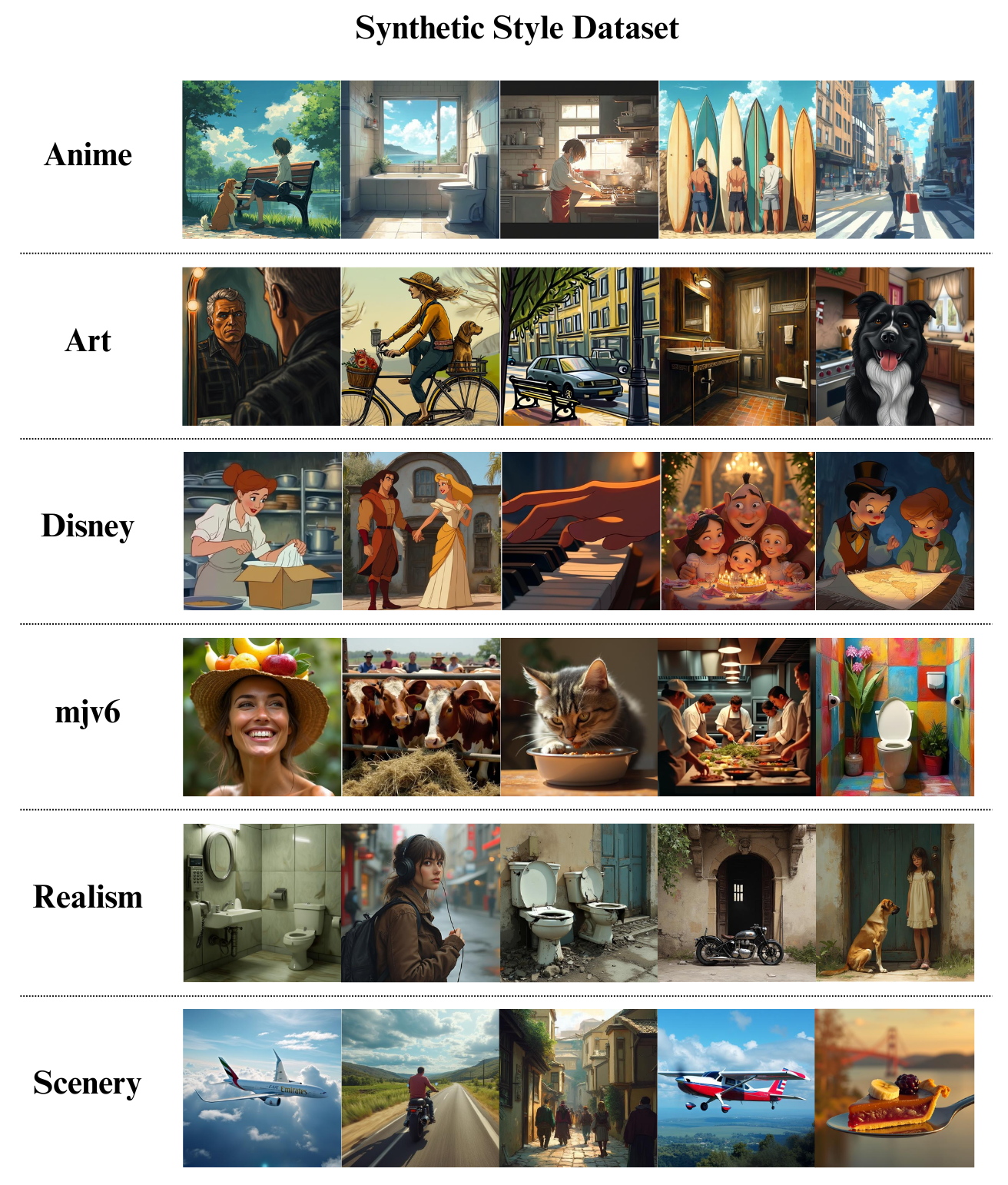}
    \caption{Additional examples of the synthetic style dataset.}
    \label{suppl_fig:synth-style}
\end{figure*}

Here we provide additional examples from the synthetic datasets: aesthetic in~\Cref{suppl_fig:synth-aes}, anatomy in~\Cref{suppl_fig:synth-anatomy}, compositionality in~\Cref{suppl_fig:synth-compo}, and style in~\Cref{suppl_fig:synth-style}.
\subsection{Prompting Strategy}
\begin{figure*}
    \centering
    \includegraphics[width=0.9\linewidth]{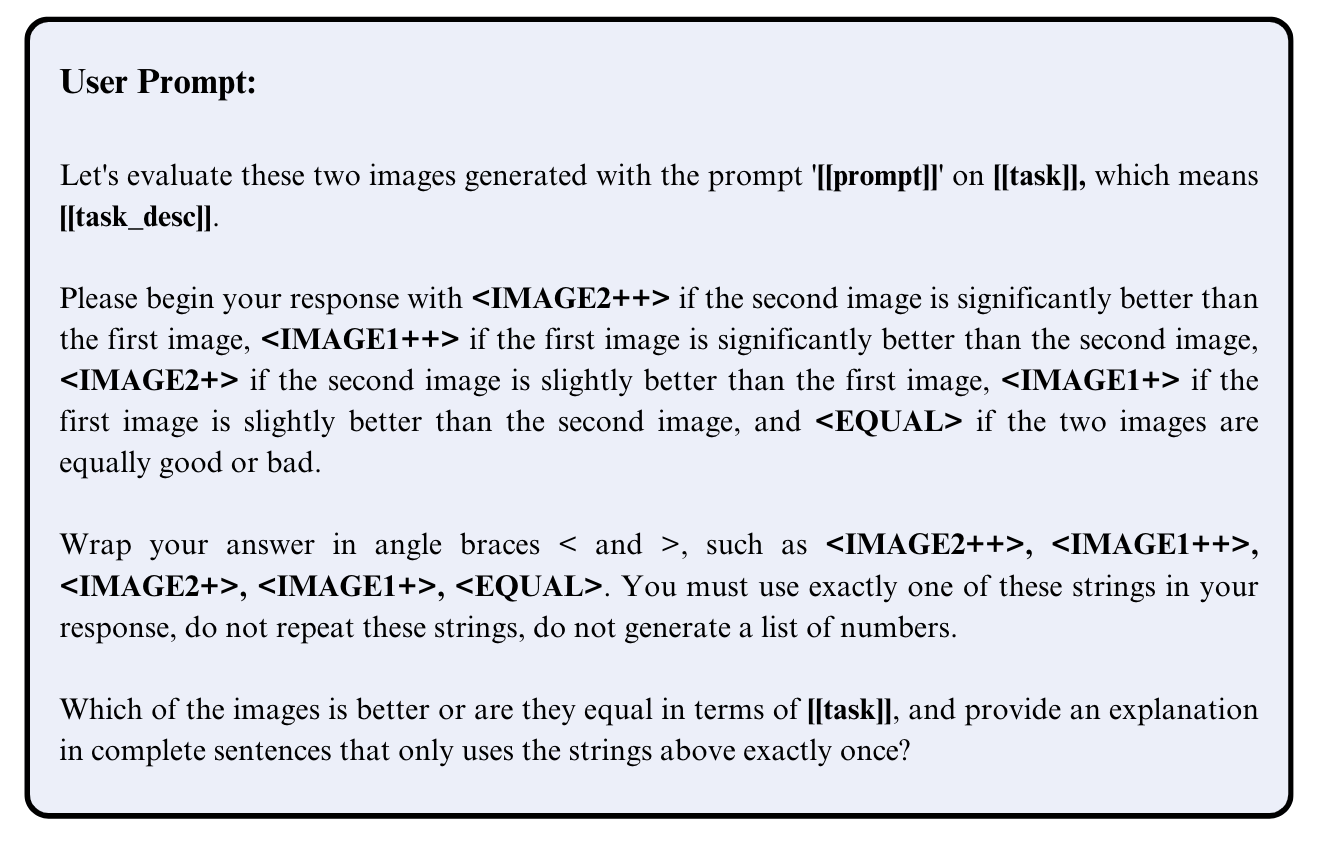}
    \caption{User prompt for multi-image evaluation.}
    \label{suppl_fig:prompt1}
\end{figure*}

\begin{figure*}
    \centering
    \includegraphics[width=0.9\linewidth]{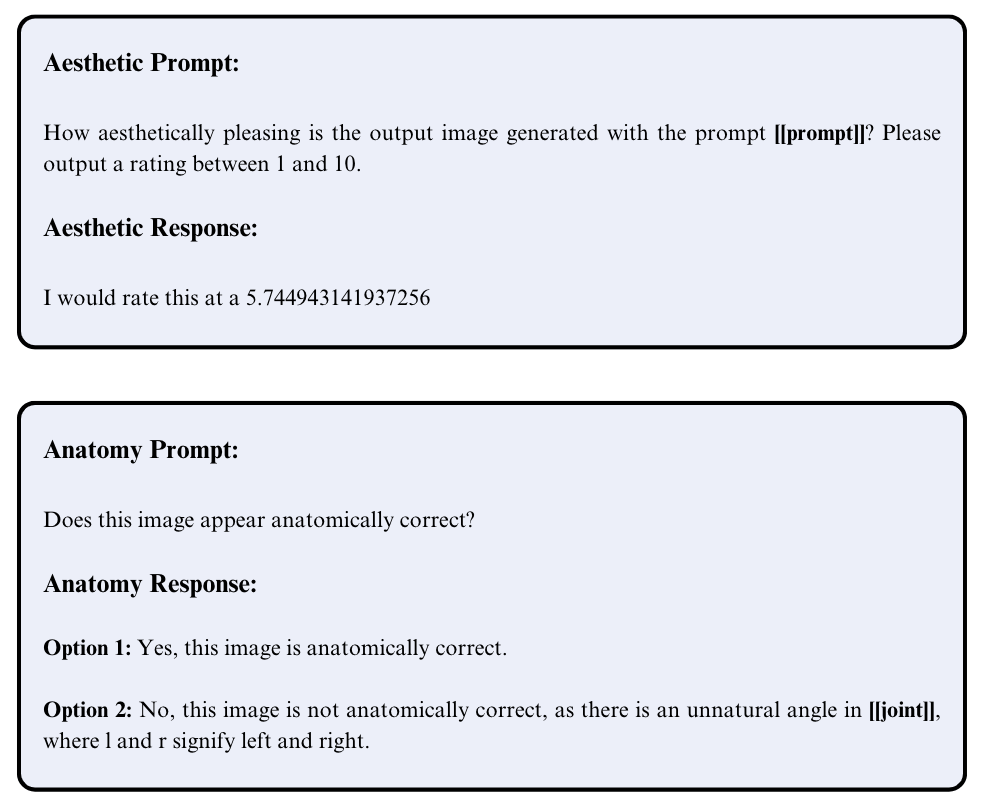}
    \caption{Prompts for aesthetic and anatomy data.}
    \label{suppl_fig:prompt2}
\end{figure*}

\begin{figure*}
    \centering
    \includegraphics[width=0.9\linewidth]{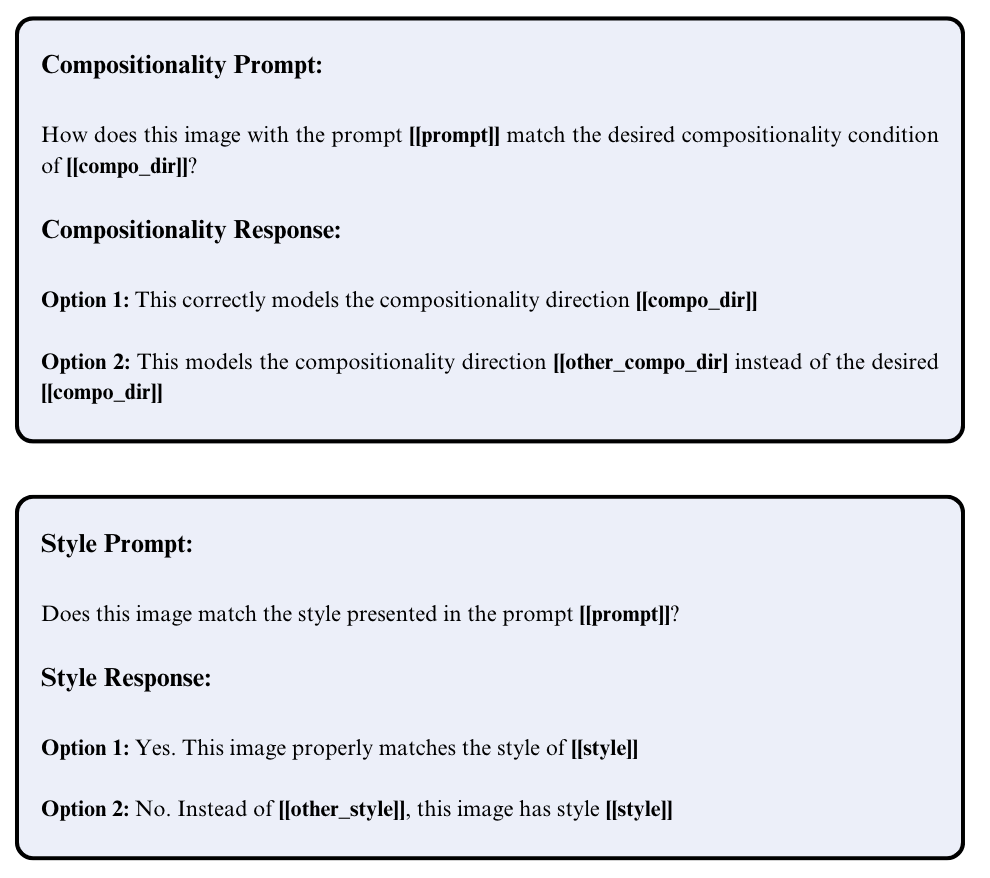}
    \caption{Prompts for compositionality and style data.}
    \label{suppl_fig:prompt3}
\end{figure*}

We present the prompts used for evaluation and training for multi-image in~\Cref{suppl_fig:prompt1}, aesthetic and anatomy in~\Cref{suppl_fig:prompt2}, and compositionality and style in~\Cref{suppl_fig:prompt3}.

\subsection{Model Training Ablations}
We sweep over a number of parameters. We sweep learning rate (5e-6 and 1e-5), number of training steps (200k, 400k, 800k), and a variety of data mixes over native LLaVA data, Commonalities and Differences (CaD) data, and our synthetic data~\cite{liu2023visual, lin2024comparison}. We also train on the SynArtifact dataset for artifacts and Img-Diff dataset for object adherence~\cite{cao2024synartifact, jiao2024imgdiff}. We evaluate the models on all validation datasets, and we find that our synthetic training paradigm on toy tasks does not generalize to the general multi-image rating task, which is expected due to the specialization of the toy datasets. We also find that training on other datasets does not transfer knowledge on a target task, with changes in accuracy being within noise. Detailed results can be found in: aesthetic toy dataset results in~\Cref{tab:aesthetic-results}, anatomy toy dataset results in~\Cref{tab:anatomy-results}, compositionality toy dataset results in~\Cref{tab:compo-results}, style toy dataset results in~\Cref{tab:style-results}, Cohen's Kappa dataset results on the overall dataset in~\Cref{tab:kappa-agreement-appendix}, and Pearson's correlation coefficient results on the overall dataset in~\Cref{tab:pearson-correlations-appendix}.

Note that models are named as\\ \texttt{[\# samples]-[\% llava data]-[\% CaD data]-[\% synthetic aesthetic data]-[\% synthetic anatomy data]-[\% SynArtifact data]-[\% synthetic compositionality data]-[\% Img-Diff data]-[\% synthetic style data]-[learning rate]}.

\newpage

\renewcommand{\arraystretch}{0.8}

\begin{table*}[h]
\centering
\begin{tabular}{lr}
\toprule
Dataset & Aesthetic Correlation (Pearson) \\
\midrule
200k-0-80-20-0-0-0-0-0-1e-5 & 0.847 \\
200k-80-0-20-0-0-0-0-0-1e-5 & 0.842 \\
200k-40-40-20-0-0-0-0-0-1e-5 & 0.840 \\
400k-40-40-10-0-0-0-10-0-lr5e-6 & 0.825 \\
200k-60-20-20-0-0-0-0-0-1e-5 & 0.818 \\
200k-50-40-10-0-0-0-0-0-1e-5 & 0.810 \\
400k-40-40-10-0-10-0-0-0-lr5e-6 & 0.799 \\
400k-40-40-10-10-0-0-0-0-lr5e-6 & 0.798 \\
800k-40-40-3-3-3-3-3-3-lr5e-6 & 0.797 \\
200k-40-40-10-0-0-0-0-10-lr5e-6 & 0.778 \\
200k-40-40-10-0-0-0-10-0-lr5e-6 & 0.773 \\
200k-40-40-10-0-0-10-0-0-lr5e-6 & 0.769 \\
400k-40-40-3-3-3-3-3-3-lr5e-6 & 0.761 \\
200k-40-40-10-10-0-0-0-0-lr5e-6 & 0.740 \\
400k-40-40-10-0-0-10-0-0-lr5e-6 & 0.719 \\
200k-40-40-3-3-3-3-3-3-lr1e-5 & 0.706 \\
200k-40-40-10-0-10-0-0-0-lr5e-6 & 0.684 \\
\textbf{gpt-4o-mini} & 0.586 \\
\textbf{gpt-4o} & 0.583 \\
400k-40-40-3-3-3-3-3-3-lr1e-5 & 0.494 \\
800k-40-40-3-3-3-3-3-3-lr1e-5 & 0.403 \\
400k-40-40-10-0-0-0-0-10-lr5e-6 & 0.379 \\
200k-40-40-3-3-3-3-3-3-lr5e-6 & 0.328 \\
\textbf{claude-3-5-sonnet-20241022} & 0.257 \\
400k-40-60-0-0-0-0-0-0-lr5e-6 & 0.178 \\
200k-40-40-0-0-0-10-10-0-lr5e-6 & 0.136 \\
400k-40-40-0-10-0-0-10-0-lr5e-6 & 0.115 \\
200k-40-40-0-0-10-10-0-0-lr5e-6 & 0.104 \\
400k-40-40-0-0-10-0-10-0-lr5e-6 & 0.100 \\
400k-40-40-0-0-10-10-0-0-lr5e-6 & 0.094 \\
200k-40-40-0-10-0-10-0-0-lr5e-6 & 0.083 \\
400k-20-80-0-0-0-0-0-0-lr5e-6 & 0.075 \\
400k-60-40-0-0-0-0-0-0-lr5e-6 & 0.074 \\
200k-40-40-0-0-0-20-0-0-lr1e-5 & 0.066 \\
200k-40-40-0-0-0-0-10-10-lr5e-6 & 0.066 \\
200k-60-40-0-0-0-0-0-0-lr5e-6 & 0.063 \\
\textbf{gemini-1.5-flash-002} & 0.037 \\
200k-40-60-0-0-0-0-0-0-lr5e-6 & 0.036 \\
400k-40-40-0-10-10-0-0-0-lr5e-6 & 0.035 \\
200k-20-80-0-0-0-0-0-0-lr5e-6 & 0.032 \\
400k-40-40-0-0-0-10-0-10-lr5e-6 & 0.023 \\
200k-40-40-0-10-0-0-10-0-lr5e-6 & 0.023 \\
200k-40-40-0-20-0-0-0-0-lr1e-5 & 0.014 \\
200k-40-40-0-0-0-0-0-20-lr1e-5 & -0.002 \\
\textbf{llava-baseline} & -0.010 \\
400k-80-20-0-0-0-0-0-0-lr5e-6 & -0.014 \\
200k-80-20-0-0-0-0-0-0-lr5e-6 & -0.014 \\
200k-40-40-0-10-10-0-0-0-lr5e-6 & -0.020 \\
400k-40-40-0-0-0-0-10-10-lr5e-6 & -0.024 \\
200k-40-40-0-0-10-0-10-0-lr5e-6 & -0.027 \\
400k-80-20-0-0-0-0-0-0-lr5e-6 & -0.033 \\
400k-40-40-0-10-0-10-0-0-lr5e-6 & -0.038 \\
400k-40-40-0-10-0-0-0-10-lr5e-6 & -0.039 \\
200k-40-40-0-0-10-0-0-10-lr5e-6 & -0.040 \\
200k-40-40-0-10-0-0-0-10-lr5e-6 & -0.041 \\
200k-40-40-0-0-0-10-0-10-lr5e-6 & -0.052 \\
\textbf{pixtral-12b} & -0.053 \\
400k-40-40-0-0-10-0-0-10-lr5e-6 & -0.122 \\
\bottomrule
\end{tabular}
\caption{Sorted correlation results for aesthetic datasets}
\label{tab:aesthetic-results}
\end{table*}
\begin{table*}[h]
\centering
\begin{tabular}{lr}
\toprule
Dataset & Anatomy Accuracy \\
\midrule
400k-40-60-0-0-0-0-0-0-lr5e-6 & 57.6\% \\
400k-60-40-0-0-0-0-0-0-lr5e-6 & 56.6\% \\
200k-40-60-0-0-0-0-0-0-lr5e-6 & 56.2\% \\
400k-80-20-0-0-0-0-0-0-lr5e-6 & 55.5\% \\
200k-40-40-0-10-0-0-10-0-lr5e-6 & 55.4\% \\
400k-20-80-0-0-0-0-0-0-lr5e-6 & 55.2\% \\
800k-40-40-3-3-3-3-3-3-lr5e-6 & 54.0\% \\
200k-40-40-10-10-0-0-0-0-lr5e-6 & 53.2\% \\
200k-80-0-0-20-0-0-0-0-1e-5 & 53.0\% \\
400k-40-40-0-10-10-0-0-0-lr5e-6 & 52.8\% \\
400k-40-40-0-10-0-0-0-10-lr5e-6 & 52.8\% \\
400k-40-40-10-0-0-10-0-0-lr5e-6 & 52.6\% \\
200k-60-40-0-0-0-0-0-0-lr5e-6 & 52.5\% \\
400k-40-40-10-0-10-0-0-0-lr5e-6 & 52.4\% \\
200k-40-40-0-10-10-0-0-0-lr5e-6 & 52.2\% \\
200k-40-40-3-3-3-3-3-3-lr1e-5 & 52.2\% \\
\textbf{claude-3-5-sonnet-20241022} & 52.0\% \\
200k-40-40-0-20-0-0-0-0-1e-5 & 52.0\% \\
200k-60-20-0-20-0-0-0-0-1e-5 & 52.0\% \\
200k-40-40-10-0-0-10-0-0-lr5e-6 & 52.0\% \\
400k-40-40-3-3-3-3-3-3-lr1e-5 & 52.0\% \\
400k-40-40-10-10-0-0-0-0-lr5e-6 & 51.8\% \\
200k-40-40-0-0-0-0-0-20-lr1e-5 & 51.8\% \\
200k-40-40-10-0-10-0-0-0-lr5e-6 & 51.8\% \\
200k-80-20-0-0-0-0-0-0-lr5e-6 & 51.7\% \\
\textbf{gpt-4o} & 51.6\% \\
200k-40-40-20-0-0-0-0-0-1e-5 & 51.6\% \\
200k-40-40-0-10-0-10-0-0-lr5e-6 & 51.4\% \\
200k-50-40-0-10-0-0-0-0-1e-5 & 51.4\% \\
400k-40-40-0-0-10-0-10-0-lr5e-6 & 51.2\% \\
\textbf{pixtral-12b} & 51.2\% \\
200k-40-40-0-10-0-0-0-10-lr5e-6 & 51.2\% \\
400k-40-40-0-0-0-0-10-10-lr5e-6 & 51.0\% \\
200k-0-80-0-20-0-0-0-0-1e-5 & 51.0\% \\
800k-40-40-3-3-3-3-3-3-lr1e-5 & 51.0\% \\
400k-40-40-10-0-0-0-10-0-lr5e-6 & 50.4\% \\
400k-40-40-0-0-0-10-10-0-lr5e-6 & 50.2\% \\
200k-40-40-0-0-0-10-10-0-lr5e-6 & 50.1\% \\
400k-40-40-10-0-0-0-0-10-lr5e-6 & 50.0\% \\
200k-40-40-3-3-3-3-3-3-lr5e-6 & 50.0\% \\
400k-40-40-0-0-10-0-0-10-lr5e-6 & 50.0\% \\
\textbf{llava-baseline} & 50.0\% \\
400k-40-40-0-0-0-10-0-10-lr5e-6 & 49.8\% \\
200k-40-40-0-0-10-0-0-10-lr5e-6 & 49.8\% \\
400k-40-40-0-10-0-0-10-0-lr5e-6 & 49.6\% \\
400k-40-40-3-3-3-3-3-3-lr5e-6 & 49.4\% \\
200k-40-40-10-0-0-0-0-10-lr5e-6 & 49.0 \% \\
200k-20-80-0-0-0-0-0-0-lr5e-6 & 48.6\% \\
400k-40-40-0-0-10-10-0-0-lr5e-6 & 48.5\% \\
200k-40-40-0-0-0-0-10-10-lr5e-6 & 48.4\% \\
200k-40-40-0-0-10-0-10-0-lr5e-6 & 48.4\% \\
200k-40-40-0-0-0-10-0-10-lr5e-6 & 48.0\% \\
\textbf{gemini-1.5-flash-002} & 47.8\% \\
200k-40-40-0-0-0-20-0-0-1e-5 & 47.8\% \\
\textbf{gpt-4o-mini} & 47.7\% \\
200k-40-40-10-0-0-0-10-0-lr5e-6 & 47.3\% \\
400k-40-40-0-10-0-10-0-0-lr5e-6 & 47.0\% \\
200k-40-40-0-0-10-10-0-0-lr5e-6 & 46.1\% \\
\bottomrule
\end{tabular}
\caption{Accuracy results for anatomy datasets (sorted by accuracy)}
\label{tab:anatomy-results}
\end{table*}

\begin{table*}[h]
\centering
\begin{tabular}{lr}
\toprule
Dataset & Compo Accuracy \\
\midrule
200k-80-0-0-0-0-20-0-0-lr1e-5 & 91.2\% \\
200k-40-40-0-10-0-10-0-0-lr5e-6 & 87.4\% \\
200k-50-40-0-0-0-10-0-0-lr1e-5 & 86.0\% \\
200k-40-40-0-0-0-10-0-10-lr5e-6 & 85.4\% \\
200k-40-40-10-0-0-10-0-0-lr5e-6 & 85.2\% \\
400k-40-40-10-0-0-10-0-0-lr5e-6 & 83.4\% \\
400k-40-40-0-0-10-10-0-0-lr5e-6 & 82.0\% \\
200k-40-40-0-0-10-10-0-0-lr5e-6 & 80.0\% \\
400k-40-40-0-0-0-10-0-10-lr5e-6 & 80.0\% \\
200k-40-40-0-0-0-10-10-0-lr5e-6 & 79.6\% \\
800k-40-40-3-3-3-3-3-3-lr5e-6 & 79.2\% \\
200k-60-20-0-0-0-20-0-0-lr1e-5 & 79.0\% \\
400k-40-40-0-0-0-10-10-0-lr5e-6 & 78.4\% \\
400k-40-40-0-10-0-10-0-0-lr5e-6 & 77.6\% \\
400k-40-40-3-3-3-3-3-3-lr1e-5 & 77.0\% \\
400k-40-40-3-3-3-3-3-3-lr5e-6 & 75.8\% \\
\textbf{gpt-4o} & 75.2\% \\
800k-40-40-3-3-3-3-3-3-lr1e-5 & 74.0\% \\
\textbf{claude-3-5-sonnet-20241022} & 71.0\% \\
200k-0-80-0-0-0-20-0-0-lr1e-5 & 68.4\% \\
200k-40-40-0-0-0-20-0-0-lr1e-5 & 59.6\% \\
\textbf{gpt-4o-mini} & 58.6\% \\
\textbf{gemini-1.5-flash-002} & 57.4\% \\
400k-40-40-0-0-0-0-10-10-lr5e-6 & 53.2\% \\
200k-40-40-0-10-0-0-0-10-lr5e-6 & 52.4\% \\
200k-40-40-10-0-0-0-0-10-lr5e-6 & 52.2\% \\
200k-40-40-20-0-0-0-0-0-lr1e-5 & 52.2\% \\
200k-40-40-10-10-0-0-0-0-lr5e-6 & 52.2\% \\
400k-40-40-10-0-0-0-10-0-lr5e-6 & 52.2\% \\
400k-80-20-0-0-0-0-0-0-lr5e-6 & 52.2\% \\
400k-60-40-0-0-0-0-0-0-lr5e-6 & 52.2\% \\
400k-40-40-10-0-0-0-0-10-lr5e-6 & 52.0\% \\
400k-20-80-0-0-0-0-0-0-lr5e-6 & 52.0\% \\
400k-40-40-0-10-0-0-10-0-lr5e-6 & 52.0\% \\
400k-40-40-10-10-0-0-0-0-lr5e-6 & 52.0\% \\
\textbf{llava-baseline} & 52.0\% \\
400k-40-40-10-0-10-0-0-0-lr5e-6 & 52.0\% \\
200k-40-40-0-20-0-0-0-0-lr1e-5 & 52.0\% \\
200k-40-40-10-0-0-0-10-0-lr5e-6 & 52.0\% \\
200k-40-40-0-10-10-0-0-0-lr5e-6 & 52.0\% \\
200k-60-40-0-0-0-0-0-0-lr5e-6 & 52.0\% \\
200k-40-60-0-0-0-0-0-0-lr5e-6 & 52.0\% \\
200k-40-40-0-10-0-0-10-0-lr5e-6 & 51.8\% \\
200k-20-80-0-0-0-0-0-0-lr5e-6 & 51.8\% \\
400k-40-40-0-0-10-0-10-0-lr5e-6 & 51.8\% \\
200k-40-40-0-0-0-0-10-10-lr5e-6 & 51.8\% \\
400k-40-40-0-10-10-0-0-0-lr5e-6 & 51.8\% \\
400k-40-60-0-0-0-0-0-0-lr5e-6 & 51.8\% \\
200k-40-40-0-0-10-0-0-10-lr5e-6 & 51.8\% \\
200k-40-40-0-0-10-0-10-0-lr5e-6 & 51.6\% \\
200k-40-40-10-0-10-0-0-0-lr5e-6 & 51.6\% \\
200k-40-40-0-0-0-0-0-20-lr1e-5 & 51.4\% \\
200k-80-20-0-0-0-0-0-0-lr5e-6 & 51.4\% \\
400k-40-40-0-0-10-0-0-10-lr5e-6 & 51.0\% \\
200k-40-40-3-3-3-3-3-3-lr1e-5 & 49.4\% \\
200k-40-40-3-3-3-3-3-3-lr5e-6 & 48.4\% \\
400k-40-40-0-10-0-0-0-10-lr5e-6 & 48.2\% \\
\textbf{pixtral-12b} & 46.6\% \\
\bottomrule
\end{tabular}
\caption{Accuracy results for compo datasets (sorted by accuracy)}
\label{tab:compo-results}
\end{table*}

\begin{table*}[h]
\centering
\begin{tabular}{lr}
\toprule
Dataset & Style Accuracy \\
\midrule
200k-0-80-0-0-0-0-0-20-lr1e-5 & 81.8\% \\
800k-40-40-3-3-3-3-3-3-lr1e-5 & 80.2\% \\
200k-80-0-0-0-0-0-0-20-lr1e-5 & 79.7\% \\
200k-40-40-0-0-0-0-0-20-lr1e-5 & 79.5\% \\
400k-40-40-0-0-0-0-10-10-lr5e-6 & 78.8\% \\
200k-60-20-0-0-0-0-0-20-lr1e-5 & 78.8\% \\
400k-40-40-10-0-0-0-0-10-lr5e-6 & 78.7\% \\
200k-50-40-0-0-0-0-0-10-lr1e-5 & 78.5\% \\
200k-40-40-0-0-10-0-0-10-lr5e-6 & 77.8\% \\
400k-40-40-0-10-0-0-0-10-lr5e-6 & 77.5\% \\
200k-40-40-10-0-0-0-0-10-lr5e-6 & 77.3\% \\
400k-40-40-0-0-0-10-0-10-lr5e-6 & 77.3\% \\
800k-40-40-3-3-3-3-3-3-lr5e-6 & 77.3\% \\
400k-40-40-0-0-10-0-0-10-lr5e-6 & 77.2\% \\
200k-40-40-0-10-0-0-0-10-lr5e-6 & 77.2\% \\
200k-40-40-0-0-0-10-0-10-lr5e-6 & 77.2\% \\
400k-40-40-3-3-3-3-3-3-lr1e-5 & 75.5\% \\
400k-40-40-3-3-3-3-3-3-lr5e-6 & 75.5\% \\
200k-40-40-0-0-0-0-10-10-lr5e-6 & 75.5\% \\
200k-40-40-3-3-3-3-3-3-lr5e-6 & 74.8\% \\
200k-40-40-3-3-3-3-3-3-lr1e-5 & 73.5\% \\
\textbf{gemini-1.5-flash-002} & 62.7\% \\
200k-40-40-10-0-0-0-10-0-lr5e-6 & 62.1\% \\
\textbf{gpt-4o} & 61.8\% \\
\textbf{gpt-4o-mini} & 61.0\% \\
\textbf{claude-3-5-sonnet-20241022} & 60.4\% \\
400k-40-40-0-0-10-0-10-0-lr5e-6 & 60.2\% \\
200k-40-40-10-0-10-0-0-0-lr5e-6 & 60.0\% \\
200k-40-40-0-0-10-10-0-0-lr5e-6 & 59.7\% \\
400k-20-80-0-0-0-0-0-0-lr5e-6 & 59.6\% \\
200k-40-40-0-0-0-10-10-0-lr5e-6 & 59.3\% \\
200k-40-40-20-0-0-0-0-0-1e-5 & 59.2\% \\
200k-40-40-10-0-0-10-0-0-lr5e-6 & 59.2\% \\
400k-40-60-0-0-0-0-0-0-lr5e-6 & 58.8\% \\
400k-80-20-0-0-0-0-0-0-lr5e-6 & 58.4\% \\
400k-40-40-10-0-0-10-0-0-lr5e-6 & 58.2\% \\
200k-40-40-0-0-10-0-10-0-lr5e-6 & 58.1\% \\
200k-80-20-0-0-0-0-0-0-lr5e-6 & 58.0\% \\
400k-40-40-10-0-0-0-10-0-lr5e-6 & 57.9\% \\
200k-40-40-0-20-0-0-0-0-1e-5 & 57.5\% \\
400k-40-40-0-0-10-10-0-0-lr5e-6 & 57.5\% \\
200k-60-40-0-0-0-0-0-0-lr5e-6 & 57.5\% \\
200k-40-40-10-10-0-0-0-0-lr5e-6 & 57.3\% \\
\textbf{llava-baseline} & 57.2\% \\
400k-40-40-0-0-0-10-10-0-lr5e-6 & 57.2\% \\
400k-40-40-0-10-0-10-0-0-lr5e-6 & 57.0\% \\
400k-40-40-10-0-10-0-0-0-lr5e-6 & 56.8\% \\
200k-40-60-0-0-0-0-0-0-lr5e-6 & 56.8\% \\
400k-40-40-10-10-0-0-0-0-lr5e-6 & 56.7\% \\
400k-40-40-0-10-0-0-10-0-lr5e-6 & 56.7\% \\
400k-60-40-0-0-0-0-0-0-lr5e-6 & 56.1\% \\
200k-20-80-0-0-0-0-0-0-lr5e-6 & 55.9\% \\
\textbf{pixtral-12b} & 55.7\% \\
200k-40-40-0-10-10-0-0-0-lr5e-6 & 55.5\% \\
200k-40-40-0-10-0-0-10-0-lr5e-6 & 55.3\% \\
400k-40-40-0-10-10-0-0-0-lr5e-6 & 55.2\% \\
200k-40-40-0-0-0-20-0-0-1e-5 & 54.3\% \\
200k-40-40-0-10-0-10-0-0-lr5e-6 & 53.8\% \\
\bottomrule
\end{tabular}
\caption{Accuracy results for style datasets (sorted by accuracy)}
\label{tab:style-results}
\end{table*}

\begin{table*}[h]
\centering
\begin{tabular}{lccccccc}
\toprule
Dataset & Aesthetic & Composition & Style & Artifacts & Anatomy & Objects & Overall \\
\midrule
\textbf{gpt4o} & \textbf{0.127} & \textbf{0.147} & \textbf{0.110} & \textbf{0.117} & \textbf{0.152} & \textbf{0.229} & \textbf{0.176} \\
\textbf{pickscore} & \textbf{-} & \textbf{-} & \textbf{-} & \textbf{-} & \textbf{-} & \textbf{-} & \textbf{0.164} \\
\textbf{claude} & \textbf{0.119} & \textbf{0.114} & \textbf{0.095} & \textbf{0.104} & \textbf{0.147} & \textbf{0.207} & \textbf{0.140} \\
\textbf{imagereward} & \textbf{-} & \textbf{-} & \textbf{-} & \textbf{-} & \textbf{-} & \textbf{-} & \textbf{0.128} \\
\textbf{vqascore} & \textbf{-} & \textbf{-} & \textbf{-} & \textbf{-} & \textbf{-} & \textbf{-} & \textbf{0.110} \\
\textbf{gemini-1.5-flash} & \textbf{0.051} & \textbf{0.047} & \textbf{0.101} & \textbf{0.062} & \textbf{0.075} & \textbf{0.138} & \textbf{0.071} \\
\textbf{gpt4o-mini} & \textbf{0.042} & \textbf{0.091} & \textbf{0.055} & \textbf{0.056} & \textbf{0.046} & \textbf{0.170} & \textbf{0.063} \\
\textbf{pixtral} & \textbf{0.030} & \textbf{0.023} & \textbf{0.028} & \textbf{0.013} & \textbf{0.017} & \textbf{0.041} & \textbf{0.045} \\
400k-40-40-0-0-10-10-0-0-lr5e-6 & 0.021 & 0.004 & 0.030 & 0.011 & 0.010 & -0.021 & 0.043 \\
400k-40-40-0-0-10-0-0-10-lr5e-6 & 0.019 & -0.010 & 0.019 & 0.005 & 0.026 & -0.024 & 0.030 \\
400k-40-40-0-10-0-0-0-10-lr5e-6 & 0.003 & 0.005 & 0.021 & 0.012 & 0.016 & 0.013 & 0.030 \\
200k-40-40-0-0-0-0-10-10-lr5e-6 & 0.026 & 0.001 & -0.009 & -0.009 & 0.003 & -0.012 & 0.023 \\
400k-40-40-10-0-0-10-0-0-lr5e-6 & 0.020 & 0.016 & 0.031 & 0.019 & 0.024 & 0.012 & 0.023 \\
400k-40-40-0-0-0-10-10-0-lr5e-6 & 0.009 & -0.001 & 0.025 & 0.023 & -0.020 & -0.003 & 0.021 \\
400k-40-40-10-0-0-0-10-0-lr5e-6 & 0.015 & 0.003 & 0.015 & 0.011 & -0.010 & -0.001 & 0.021 \\
200k-40-40-10-0-0-10-0-0-lr5e-6 & 0.015 & 0.015 & 0.050 & 0.029 & -0.007 & 0.016 & 0.020 \\
200k-40-40-3-3-3-3-3-3-lr5e-6 & 0.019 & 0.002 & 0.009 & 0.030 & 0.016 & -0.009 & 0.018 \\
200k-20-80-0-0-0-0-0-0-lr5e-6 & 0.026 & -0.005 & -0.003 & -0.025 & 0.022 & -0.025 & 0.018 \\
400k-40-40-0-10-0-10-0-0-lr5e-6 & -0.006 & 0.028 & 0.014 & 0.025 & 0.019 & 0.016 & 0.017 \\
400k-40-40-10-10-0-0-0-0-lr5e-6 & -0.020 & 0.007 & -0.000 & 0.008 & 0.003 & -0.003 & 0.016 \\
200k-40-40-0-0-0-10-10-0-lr5e-6 & -0.008 & -0.005 & 0.024 & 0.016 & 0.015 & -0.024 & 0.016 \\
200k-40-40-0-10-0-0-0-10-lr5e-6 & 0.013 & 0.006 & 0.018 & -0.012 & 0.008 & -0.012 & 0.012 \\
200k-40-40-0-10-0-0-10-0-lr5e-6 & -0.003 & -0.019 & -0.004 & 0.013 & 0.073 & -0.016 & 0.010 \\
400k-40-40-3-3-3-3-3-3-lr5e-6 & 0.010 & -0.005 & -0.012 & 0.007 & 0.005 & -0.008 & 0.009 \\
400k-40-40-3-3-3-3-3-3-lr1e-5 & 0.017 & 0.015 & 0.046 & 0.003 & -0.002 & 0.011 & 0.008 \\
200k-40-40-0-0-0-10-0-10-lr5e-6 & -0.015 & -0.022 & 0.009 & 0.019 & 0.038 & -0.021 & 0.008 \\
400k-40-60-0-0-0-0-0-0-lr5e-6 & 0.015 & -0.021 & -0.012 & 0.017 & 0.019 & -0.008 & 0.008 \\
400k-40-40-10-0-10-0-0-0-lr5e-6 & 0.004 & -0.006 & -0.001 & 0.008 & 0.021 & -0.000 & 0.008 \\
200k-40-40-10-0-0-0-0-10-lr5e-6 & -0.007 & -0.005 & -0.007 & 0.007 & 0.017 & -0.019 & 0.006 \\
400k-40-40-0-0-0-10-0-10-lr5e-6 & -0.002 & -0.008 & 0.021 & 0.022 & -0.004 & 0.019 & 0.006 \\
400k-40-40-10-0-0-0-0-10-lr5e-6 & 0.020 & 0.007 & 0.017 & 0.029 & 0.033 & -0.005 & 0.006 \\
400k-80-20-0-0-0-0-0-0-lr5e-6 & -0.001 & 0.000 & 0.011 & 0.007 & -0.012 & -0.028 & 0.004 \\
400k-40-40-3-3-3-3-3-3-lr5e-6 & 0.014 & -0.008 & 0.019 & 0.004 & 0.020 & 0.004 & 0.004 \\
400k-40-40-0-10-0-0-10-0-lr5e-6 & 0.002 & -0.001 & 0.013 & 0.020 & 0.003 & -0.013 & 0.003 \\
200k-40-40-0-0-10-0-0-10-lr5e-6 & 0.012 & 0.019 & 0.008 & 0.017 & 0.018 & -0.001 & 0.003 \\
\textbf{llava-baseline} & -0.001 & 0.019 & -0.007 & -0.000 & -0.004 & -0.006 & 0.002 \\
200k-40-40-20-0-0-0-0-0-lr1e-5 & -0.007 & -0.009 & -0.009 & 0.032 & 0.010 & -0.003 & 0.002 \\
400k-60-40-0-0-0-0-0-0-lr5e-6 & 0.023 & -0.004 & 0.022 & 0.011 & 0.034 & -0.030 & 0.001 \\
400k-40-40-0-10-10-0-0-0-lr5e-6 & 0.018 & -0.007 & -0.005 & -0.022 & 0.017 & -0.005 & 0.000 \\
800k-40-40-3-3-3-3-3-3-lr5e-6 & 0.017 & 0.000 & 0.007 & -0.004 & 0.030 & 0.005 & -0.001 \\
200k-40-40-0-10-0-10-0-0-lr5e-6 & 0.016 & 0.003 & -0.028 & -0.002 & 0.021 & 0.003 & -0.001 \\
200k-40-40-0-0-0-20-0-0-lr1e-5 & 0.026 & -0.015 & 0.014 & 0.013 & -0.001 & -0.000 & -0.001 \\
200k-80-20-0-0-0-0-0-0-lr5e-6 & 0.022 & 0.019 & 0.005 & 0.032 & 0.017 & 0.001 & -0.002 \\
200k-40-40-0-10-10-0-0-0-lr5e-6 & 0.011 & -0.003 & 0.018 & -0.031 & 0.000 & 0.004 & -0.002 \\
200k-40-60-0-0-0-0-0-0-lr5e-6 & 0.013 & 0.001 & 0.020 & -0.015 & 0.012 & 0.014 & -0.002 \\
200k-40-40-10-0-0-0-10-0-lr5e-6 & 0.004 & -0.009 & 0.017 & 0.006 & -0.005 & -0.028 & -0.002 \\
800k-40-40-3-3-3-3-3-3-lr1e-5 & -0.016 & -0.030 & 0.014 & 0.013 & 0.007 & -0.018 & -0.003 \\
200k-40-40-10-10-0-0-0-0-lr5e-6 & -0.003 & -0.007 & -0.013 & 0.009 & 0.001 & 0.018 & -0.006 \\
200k-40-40-10-0-10-0-0-0-lr5e-6 & 0.007 & -0.020 & 0.010 & 0.013 & 0.013 & -0.017 & -0.006 \\
200k-40-40-0-20-0-0-0-0-lr1e-5 & 0.027 & -0.013 & -0.023 & 0.002 & 0.001 & -0.008 & -0.006 \\
200k-40-40-0-0-0-0-0-20-lr1e-5 & 0.006 & -0.017 & -0.003 & 0.014 & 0.006 & -0.012 & -0.007 \\
400k-20-80-0-0-0-0-0-0-lr5e-6 & -0.015 & -0.030 & -0.003 & 0.021 & 0.004 & -0.041 & -0.007 \\
200k-40-40-0-0-10-0-10-0-lr5e-6 & -0.015 & 0.008 & 0.005 & 0.001 & -0.004 & -0.012 & -0.009 \\
400k-40-40-0-0-0-0-10-10-lr5e-6 & 0.022 & 0.000 & 0.039 & 0.004 & -0.008 & 0.017 & -0.009 \\
400k-40-40-0-0-10-0-10-0-lr5e-6 & -0.020 & 0.003 & 0.023 & -0.003 & 0.005 & -0.003 & -0.010 \\
200k-40-40-3-3-3-3-3-3-lr1e-5 & -0.024 & -0.003 & -0.009 & 0.003 & 0.009 & 0.008 & -0.013 \\
200k-60-40-0-0-0-0-0-0-lr5e-6 & -0.019 & -0.003 & -0.012 & -0.011 & 0.008 & 0.002 & -0.015 \\
200k-40-40-0-0-10-10-0-0-lr5e-6 & 0.014 & -0.019 & 0.025 & 0.047 & -0.007 & -0.014 & -0.030 \\
\bottomrule
\end{tabular}
\caption{Cohen's Kappa agreement between human and LLM ratings across different aspects (sorted by Overall)}
\label{tab:kappa-agreement-appendix}
\end{table*}

\begin{table*}[h]
\centering
\begin{tabular}{lccccccc}
\toprule
Dataset & Aesthetic & Composition & Style & Artifacts & Anatomy & Objects & Overall \\
\midrule
\textbf{pickscore} & \textbf{-} & \textbf{-} & \textbf{-} & \textbf{-} & \textbf{-} & \textbf{-} & \textbf{0.498} \\
\textbf{gpt4o} & \textbf{0.376} & \textbf{0.467} & \textbf{0.337} & \textbf{0.349} & \textbf{0.398} & \textbf{0.501} & \textbf{0.461} \\
\textbf{claude} & \textbf{0.338} & \textbf{0.397} & \textbf{0.336} & \textbf{0.361} & \textbf{0.390} & \textbf{0.467} & \textbf{0.407} \\
\textbf{imagereward} & \textbf{-} & \textbf{-} & \textbf{-} & \textbf{-} & \textbf{-} & \textbf{-} & \textbf{0.398} \\
\textbf{vqascore} & \textbf{-} & \textbf{-} & \textbf{-} & \textbf{-} & \textbf{-} & \textbf{-} & \textbf{0.356} \\
\textbf{gemini-1.5-flash} & \textbf{0.260} & \textbf{0.339} & \textbf{0.282} & \textbf{0.307} & \textbf{0.337} & \textbf{0.360} & \textbf{0.362} \\
\textbf{gpt4o-mini} & \textbf{0.211} & \textbf{0.259} & \textbf{0.152} & \textbf{0.175} & \textbf{0.167} & \textbf{0.350} & \textbf{0.286} \\
\textbf{pixtral} & \textbf{0.052} & \textbf{0.119} & \textbf{0.080} & \textbf{0.053} & \textbf{0.074} & \textbf{0.154} & \textbf{0.140} \\
200k-40-40-0-10-10-0-0-0-lr5e-6 & 0.027 & -0.010 & -0.004 & -0.040 & -0.012 & 0.012 & 0.072 \\
400k-40-40-0-0-10-10-0-0-lr5e-6 & -0.009 & -0.069 & 0.079 & 0.023 & 0.024 & 0.033 & 0.059 \\
400k-40-40-0-10-0-0-0-10-lr5e-6 & -0.018 & 0.074 & 0.051 & 0.013 & 0.032 & 0.048 & 0.043 \\
200k-40-40-0-0-0-0-10-10-lr5e-6 & 0.029 & 0.020 & -0.020 & -0.030 & -0.033 & 0.015 & 0.040 \\
400k-40-40-10-0-10-0-0-0-lr5e-6 & 0.008 & -0.038 & -0.056 & 0.022 & 0.059 & -0.010 & 0.036 \\
200k-40-40-0-20-0-0-0-0-lr1e-5 & -0.005 & 0.001 & 0.030 & 0.029 & -0.009 & -0.057 & 0.031 \\
400k-40-40-10-0-0-0-10-0-lr5e-6 & 0.030 & -0.007 & -0.006 & 0.016 & -0.029 & -0.003 & 0.026 \\
200k-40-40-3-3-3-3-3-3-lr5e-6 & 0.060 & 0.015 & -0.062 & 0.039 & -0.031 & 0.025 & 0.023 \\
400k-40-40-10-0-0-10-0-0-lr5e-6 & 0.026 & 0.054 & 0.025 & 0.018 & 0.049 & -0.001 & 0.016 \\
400k-40-40-10-0-0-0-0-10-lr5e-6 & 0.031 & 0.025 & 0.036 & -0.004 & 0.050 & 0.000 & 0.013 \\
400k-40-40-3-3-3-3-3-3-lr5e-6 & 0.010 & -0.015 & 0.079 & 0.030 & 0.055 & 0.023 & 0.012 \\
200k-40-40-0-0-0-0-0-20-lr1e-5 & -0.024 & 0.008 & -0.033 & 0.005 & 0.009 & 0.021 & 0.012 \\
200k-40-40-0-10-0-0-0-10-lr5e-6 & 0.005 & 0.080 & -0.005 & -0.002 & -0.043 & 0.042 & 0.010 \\
400k-40-40-0-10-0-10-0-0-lr5e-6 & -0.018 & 0.032 & 0.005 & 0.023 & 0.046 & 0.036 & 0.006 \\
400k-40-40-0-0-0-0-10-10-lr5e-6 & 0.058 & -0.008 & 0.002 & 0.043 & -0.014 & 0.050 & 0.004 \\
400k-80-20-0-0-0-0-0-0-lr5e-6 & 0.013 & 0.015 & 0.030 & 0.006 & 0.007 & 0.019 & 0.000 \\
200k-80-20-0-0-0-0-0-0-lr5e-6 & 0.022 & -0.008 & -0.049 & -0.005 & -0.002 & -0.007 & -0.001 \\
200k-40-40-0-0-0-10-10-0-lr5e-6 & 0.010 & -0.045 & -0.077 & -0.024 & -0.032 & -0.006 & -0.003 \\
200k-40-40-0-0-0-20-0-0-lr1e-5 & 0.002 & 0.001 & 0.018 & 0.055 & 0.025 & -0.004 & -0.006 \\
\textbf{llava-baseline} & -0.032 & 0.020 & -0.042 & 0.028 & 0.019 & -0.004 & -0.007 \\
400k-60-40-0-0-0-0-0-0-lr5e-6 & 0.029 & 0.017 & 0.007 & -0.014 & 0.016 & -0.040 & -0.008 \\
400k-40-40-3-3-3-3-3-3-lr1e-5 & 0.022 & 0.011 & 0.118 & 0.008 & 0.007 & -0.037 & -0.008 \\
200k-40-40-10-0-10-0-0-0-lr5e-6 & 0.016 & -0.019 & 0.007 & -0.004 & 0.038 & 0.017 & -0.010 \\
200k-40-40-0-0-10-0-0-10-lr5e-6 & 0.017 & 0.038 & 0.077 & 0.030 & 0.030 & 0.039 & -0.010 \\
200k-40-40-10-0-0-10-0-0-lr5e-6 & -0.036 & -0.014 & 0.077 & 0.018 & -0.004 & -0.018 & -0.011 \\
200k-40-40-10-0-0-0-10-0-lr5e-6 & -0.041 & 0.024 & 0.059 & 0.029 & -0.045 & 0.005 & -0.012 \\
200k-40-40-20-0-0-0-0-0-lr1e-5 & 0.003 & -0.034 & -0.036 & -0.042 & 0.018 & 0.056 & -0.012 \\
400k-40-40-0-10-10-0-0-0-lr5e-6 & -0.024 & 0.008 & 0.049 & 0.011 & -0.004 & 0.027 & -0.018 \\
800k-40-40-3-3-3-3-3-3-lr1e-5 & -0.018 & -0.025 & -0.025 & 0.016 & -0.021 & -0.014 & -0.019 \\
200k-20-80-0-0-0-0-0-0-lr5e-6 & 0.043 & -0.025 & 0.012 & -0.010 & -0.015 & -0.012 & -0.020 \\
800k-40-40-3-3-3-3-3-3-lr5e-6 & 0.027 & 0.033 & -0.043 & -0.013 & 0.041 & 0.010 & -0.021 \\
200k-40-40-0-10-0-10-0-0-lr5e-6 & 0.029 & -0.013 & -0.011 & -0.024 & 0.012 & -0.014 & -0.022 \\
200k-40-40-10-0-0-0-0-10-lr5e-6 & 0.019 & 0.012 & -0.048 & -0.042 & 0.004 & -0.005 & -0.023 \\
400k-40-40-0-0-0-10-10-0-lr5e-6 & -0.006 & 0.005 & -0.055 & 0.000 & -0.001 & 0.017 & -0.024 \\
400k-40-40-0-0-10-0-0-10-lr5e-6 & 0.044 & -0.005 & 0.005 & 0.017 & 0.028 & 0.001 & -0.024 \\
200k-40-40-0-0-10-0-10-0-lr5e-6 & -0.034 & -0.001 & 0.066 & -0.020 & -0.006 & -0.023 & -0.024 \\
200k-40-40-0-10-0-0-10-0-lr5e-6 & -0.026 & 0.000 & -0.022 & 0.026 & 0.046 & 0.022 & -0.025 \\
400k-40-40-3-3-3-3-3-3-lr5e-6 & 0.011 & 0.000 & -0.002 & 0.036 & -0.012 & 0.001 & -0.026 \\
400k-40-40-10-10-0-0-0-0-lr5e-6 & -0.040 & 0.028 & 0.013 & 0.004 & 0.017 & -0.031 & -0.026 \\
400k-40-40-0-0-0-10-0-10-lr5e-6 & 0.019 & -0.020 & 0.037 & -0.042 & 0.006 & 0.034 & -0.028 \\
200k-40-40-10-10-0-0-0-0-lr5e-6 & 0.008 & -0.034 & -0.025 & 0.047 & -0.040 & 0.055 & -0.028 \\
400k-40-60-0-0-0-0-0-0-lr5e-6 & 0.026 & 0.004 & 0.068 & 0.011 & -0.004 & 0.019 & -0.030 \\
400k-40-40-0-0-10-0-10-0-lr5e-6 & -0.039 & 0.034 & 0.074 & -0.032 & 0.013 & -0.018 & -0.034 \\
400k-40-40-0-10-0-0-10-0-lr5e-6 & -0.008 & 0.024 & 0.090 & -0.011 & 0.003 & 0.001 & -0.036 \\
200k-40-40-0-0-10-10-0-0-lr5e-6 & 0.009 & 0.023 & 0.020 & 0.023 & 0.024 & 0.004 & -0.038 \\
200k-60-40-0-0-0-0-0-0-lr5e-6 & -0.019 & 0.006 & -0.057 & -0.026 & 0.039 & 0.026 & -0.044 \\
400k-20-80-0-0-0-0-0-0-lr5e-6 & -0.057 & -0.088 & 0.048 & -0.013 & -0.035 & -0.005 & -0.045 \\
200k-40-40-0-0-0-10-0-10-lr5e-6 & 0.029 & 0.042 & -0.009 & 0.022 & 0.042 & -0.004 & -0.049 \\
200k-40-40-3-3-3-3-3-3-lr1e-5 & -0.058 & 0.025 & 0.068 & -0.021 & 0.011 & 0.076 & -0.052 \\
200k-40-60-0-0-0-0-0-0-lr5e-6 & -0.024 & 0.006 & 0.021 & -0.019 & 0.063 & 0.088 & -0.066 \\
\bottomrule
\end{tabular}
\caption{Pearson correlation between human and LLM ratings across different aspects (sorted by Overall)}
\label{tab:pearson-correlations-appendix}
\end{table*}